\documentclass[11pt,a4paper]{article}
\usepackage[hyperref]{emnlp2020}
\usepackage{times}
\usepackage{latexsym}

\usepackage{subcaption}
\usepackage{graphicx}
\usepackage{subcaption}
\usepackage{multirow}
\usepackage{algorithm}  
\usepackage{algorithmic}
\usepackage{booktabs}
\usepackage{tabularx}
\usepackage{xcolor}
\usepackage{wrapfig, blindtext}

\newcommand{\ins}{\textsc{Insert}}
\newcommand{\del}{\textsc{Delete}}
\newcommand{\sub}{\textsc{Substitute}}

\usepackage{microtype}
\usepackage{amsthm}
\usepackage{amsmath}
\usepackage{xspace}

\renewcommand{\vec}[1]{\boldsymbol{\mathbf{#1}}}

\def\vT{\vec{T}}
\def\vx{\vec{x}}
\def\vy{\vec{y}}
\def\vz{\vec{z}}
\def\vh{\vec{h}}
\def\D{\mathcal{D}}

\def\vg{\vec{g}}
\def\vm{\vec{m}}
\def\vc{\vec{c}}

\def\va{\vec{a}}
\def\softmax{\mathrm{softmax}}
\def\vb{\vec{b}}

\def\vW{\vec{W}}

\newcommand\comment[1]{}
\def\figref#1{figure~\ref{#1}}
\def\Figref#1{Figure~\ref{#1}}

\def\tabref#1{table~\ref{#1}}
\def\Tabref#1{Table~\ref{#1}}

\def\eqref#1{(\ref{#1})}
\def\Eqref#1{Equation~\ref{#1}}

\def\Algref#1{Algorithm~\ref{#1}}

\def\cf{{\em c.f.,}\xspace}

\definecolor{mygray}{gray}{0.25}

\aclfinalcopy 

\title{Scene Graph Modification Based on Natural Language Commands}

\author{
\begin{tabular}{cccc}
\bf{Xuanli He}$^\dagger$ &  \bf{Quan Hung Tran}$^\ddagger$ & \bf{Gholamreza Haffari}$^\dagger$ & \textbf{Walter Chang}$^\ddagger$ \tabularnewline
\\
 \textbf{Trung Bui}$^\ddagger$& \textbf{Zhe Lin}$^\ddagger$ & \textbf{Franck Dernoncourt}$^\ddagger$ &\textbf{Nhan Dam}$^\dagger$\tabularnewline
 \\
 & \multicolumn{2}{c}{$^\dagger$ Monash University, Clayton, Australia} &\tabularnewline
 & \multicolumn{2}{c}{\textcolor{mygray}{\small\{xuanli.he1,~gholamreza.haffari,~nhan.dam\}@monash.edu}} &\tabularnewline
 &  \multicolumn{2}{c}{$^\ddagger$Adobe Research, San Jose, CA} &\tabularnewline
 & \multicolumn{2}{c}{\textcolor{mygray}{\small\{qtran,~wachang,~bui,~dernonco\}@adobe.com}} &\tabularnewline
\end{tabular}}

\date{}

\begin{document}
\maketitle
\begin{abstract}
Structured representations like graphs and parse trees 
play a crucial role in many Natural Language Processing
systems. In recent years, the advancements in multi-turn 
user interfaces necessitate the need for controlling and 
updating
these structured representations given new sources of
information. Although there have been many efforts
focusing on
improving the performance of the parsers that map text 
to graphs or parse trees, very few have
explored the problem of directly manipulating these
representations. In this paper,
we  explore the novel problem of graph 
modification,
where the systems need to learn how to update an existing
scene graph given a new user's command. Our novel
models based on graph-based sparse transformer and
cross attention information fusion outperform previous
systems adapted from the machine translation and graph 
generation literature. We further contribute our large
graph modification datasets to
the research community to encourage future research for this
new problem.
\end{abstract}

\section{Introduction}


Parsing text into structured semantics representation is 
one of the most long-standing and active research problems in 
NLP. Numerous parsing methods have 
been developed for many different semantic structure 
representations~\cite{chen2014fast,mrini2019rethinking,zhou2019head,clark2018semi,wang2018scene}. 
However, most
of these previous works focus on parsing a \emph{single} sentence,
while a typical human-computer interaction session or 
conversation is \emph{not single-turn}. 
A prominent example is
image search. Users usually start with short phrases
describing the main objects or topics they are looking for.
Depending on the result, the users may then \emph{modify} their
query to add more constraints or give additional 
information. 
In this case, without the modification 
capability, a static representation is not suitable to track
the changing intent of the user. We 
argue that the back-and-forth and multi-turn nature of 
human-computer interactions necessitate
the need for updating the structured representation. {Another advantage of modifying a structured representation in the interactive setting is that it makes it easier to check the consistency. For instance, it is much easier to check whether the user requests two contradicting attributes for the same object in a scene graph during the interactive search, which can be done automatically.}

In this paper, we propose the problem of \emph{scene graph}  \emph{modification}  for search.
{A scene graph  \cite{johnson2015image} is a semantic formalism which represents the desired  image as a graph
of objects with relations and attributes.}
This semantic representation has been shown to be very 
successful in retrieval 
systems~\cite{johnson2015image,schuster2015generating,vendrov2015order}. Inspired by the dialog state
tracking setting~\cite{perez2017dialog,ren2018towards}, we
consider the scene graph modification problem  as follows. 
Given an initial scene graph and a new query issued by the user, the goal is to generate a new scene graph taking
into account the original graph and the new query. 



We formulate the problem as \emph{conditional graph modification}, and create three datasets for this problem.
We propose novel encoder-decoder architectures for conditional graph modification.  
More  specifically, our graph encoder is built upon the self-attention architecture popular in state-of-the-art machine translation  models~\cite{vaswani2017attention,edunov2018understanding}, which is superior to, according to our study, Graph Convolutional Networks (GCN) ~\cite{DBLP:journals/corr/KipfW16}. 
Unique to our problem, however, is the fact that we have an open set of relation types in the graphs. Thus,
we propose a novel graph-conditioned sparse transformer, in  which the relation information is embedded directly
into the self-attention grid. 
For the decoder, 
we treat the graph modification task
as a sequence generation problem ~\cite{li2018learning,simonovsky2018graphvae,you2018graphrnn}. Furthermore, to encourage the information sharing between
the input graph and modification query, we introduce
two techniques, i.e. late feature fusion through gating and
early feature fusion through cross-attention.  We further create three datasets to evaluate our models. The first two datasets are derived from public sources: MSCOCO~\cite{lin2014microsoft} and Google Conceptual Captioning (GCC)~\cite{sharma-etal-2018-conceptual} while the last is 
collected using Amazon Mechanical Turk (MTurk).
Experiments show that our best model achieves up to 8.5\% improvement over the strong baselines on both the synthetic and user-generated data in terms of F1 score.

Our contributions are three-fold. Firstly,
we introduce the problem of scene graph modification -- an important component in multi-modal search and dialogue. 
Secondly, we 
propose a novel encoder-decoder architecture relying on
graph-conditioned transformer and cross-attention to 
tackle the problem, outperforming strong baselines which we setup for the task.
Thirdly, we introduce three datasets which can serve as evaluation benchmarks for future research.\footnote{Code and datasets are available at: https://github.com/xlhex/SceneGraphModification.git.}

\section{Data Creation}


In this section, we detail our data creation process. We
 start with information on scene graphs and a parser to generate them for the captions in two
 existing datasets, i.e. MSCOCO \cite{lin2014microsoft} and GCC \cite{sharma-etal-2018-conceptual}. 
We then describe how to generate modified scene graphs and modification queries based on these scene graphs, and 
leverage human annotators to increase and analyze data quality. 

\subsection{Scene Graphs}
\label{sec:parsing}

\begin{figure}[h]
\centering
  \includegraphics[scale=0.55]{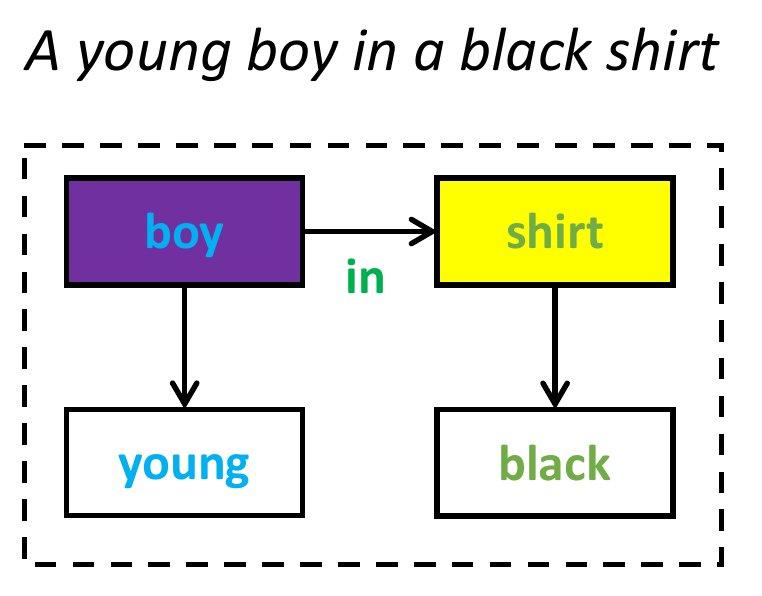}
  \caption{Example scene graph}
   \label{fig:parsing}
   \vspace{-3mm}
\end{figure}

\citet{schuster2015generating} introduce
scene graphs as semantic representations of images.
As shown in Figure~\ref{fig:parsing}, a parser will parse a sentence into a list of objects, e.g. "\textit{boy}" and "\textit{shirt}". These objects and their associated attributes and relations form a group of triplets, e.g. $\langle boy, in, shirt\rangle$, $\langle boy, attribute, young\rangle$ and $\langle shirt, attribute, black\rangle$.

Although there are several scene graphs annotated datasets for images 
\cite{krishna2017visual}, the alignments between graphs and text 
are unavailable. Moreover, image grounded scene graphs, e.g. the Visual Genome dataset~\cite{krishna2017visual}, 
also contain lots of non-salient objects and relations, while search 
queries focus more on the main objects and their connections.

The lack of 
a large-scale and high quality public dataset prompts us to create our 
own benchmark datasets. To do this, we start with the popular captioning datasets: MSCOCO \cite{lin2014microsoft} and GCC \cite{sharma-etal-2018-conceptual}. To construct scene graphs, we use an in-house scene 
graph parser to parse a random subset of MSCOCO description data and GCC captions.
The parser is built upon a dependency 
parser~\cite{dozat2016deep}, similar
to the SPICE system~\cite{anderson2016spice}. 

\subsection{Modified MSCOCO and GCC for Graph Modification}

Our first two datasets add annotations on top of the captions for MSCOCO
and GCC. The parser described in \S \ref{sec:parsing} is used to create 200k scene graphs from MSCOCO and 420k scene graphs from GCC data. Comparing the two datasets, the
graphs from MSCOCO are simpler, while the GCC graphs are 
much more complicated. According to our in-house search log, image search queries are usually short, thus the MSCOCO graphs
represents a closer match to actual search queries\footnote{Please refer to Appendix~\ref{sec:data_stat} for the statistics of our in-house search log.}, while the GCC graphs present a greater challenge to the models.

\begin{figure*}
    \centering
    \includegraphics[width=0.75\textwidth]{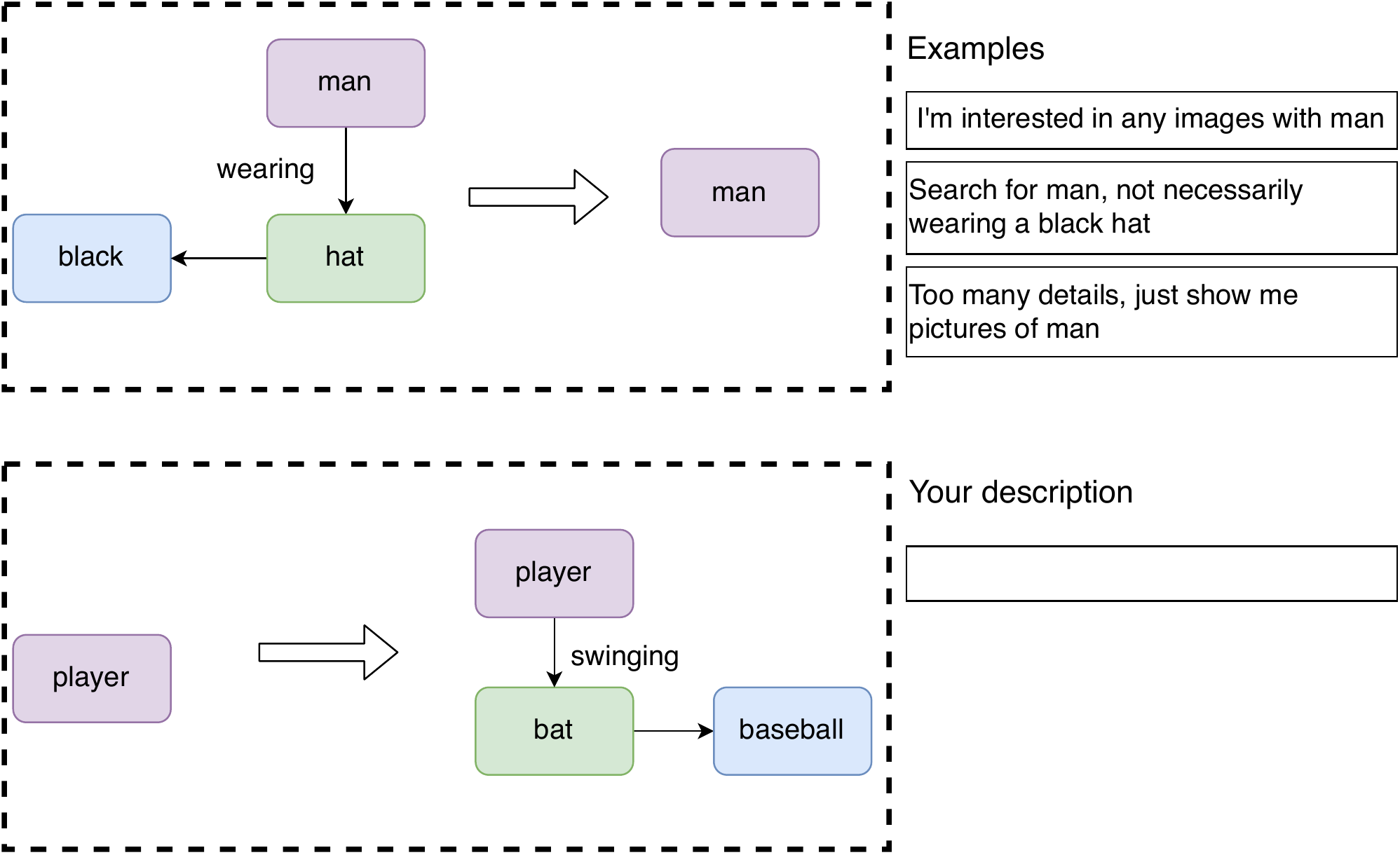}
    \caption{An interface of the crowd-sourcing stage}
    \label{fig:crowd}
\end{figure*}

Given a scene graph $\mathcal{G}$, we construct a triplet $(\vx, \vy, \vz)$, where $\vx$ is the source graph, $\vy$ indicates the modification query, and $\vz$ represents the target graph. More specifically,  we uniformly select and apply an action $\va$ from the set of all possible graph modification operations $A=\{\ins, \del, \sub\}$. The actions are applied to the graph as follows:
  \paragraph{\del.} We randomly select a node from $\mathcal{G}$ (denoting the source graph $\vx$), and then remove this node and its associated edges. The remaining nodes and edges are then the  target graph $\vz$. As for the modification query $\vy$, it is generated from a randomly selected \emph{deletion template} or by MTurk workers. These templates are based upon the Edit Me dataset~\cite{manuvinakurike2018edit}. 
    \paragraph{\ins.} We treat insertion as the inversion of deletion. Specifically, we  produce the source graph $\vx$ via a \del \ operation on $\mathcal{G}$, where the target graph $\vz$ is set to $\mathcal{G}$.  Like the deletion operator, the insertion query $\vy$ is generated by either the MTurk workers, or by templates.
    \paragraph{\sub.} We replace a randomly selected node from the source graph $\mathcal{G}$  with a semantically similar node to get the target graph. To find the new node, we make use of the AllenNLP toolkit \cite{Gardner2017AllenNLP} to get a list of candidate words based on their semantic similarity scores to the old node. More details can be found in our
  supplementary materials.

\subsection{Crowd-sourcing User Data}
\label{sec:crowd}

As described above, apart from using templates, we crowd-source more diverse and natural modification queries from MTurk. As depicted in \figref{fig:crowd}, we first show the workers an example which includes a source graph, a target graph and three acceptable modification queries. Then the workers are asked to fill in their own description for the unannotated instances. We refer to the template-based version of the datasets as ``synthetic'' while the user-generated contents as ``user-generated''.





From our preliminary trials, we notice several difficulties 
within the data collection process. Firstly, 
understanding the graphs requires some knowledge of NLP, 
thus not all MTurk workers can provide good modification queries.
Secondly, due to deletion and parser errors, we encounter
some graphs with disconnected components in the data. 
Thirdly, there are many overly 
complicated graphs which are not representative of search queries, as most of the search queries
are relatively short, with just one or two objects.
To mitigate these problems, we manually filter the data by removing  graphs with disconnected components, low-quality instances, or excessively long descriptions (i.e. more than 5 nodes). The final dataset contains 32k examples.





\begin{figure}[h]
\vspace{-4mm}
\centering
    \includegraphics[width=0.42\textwidth]{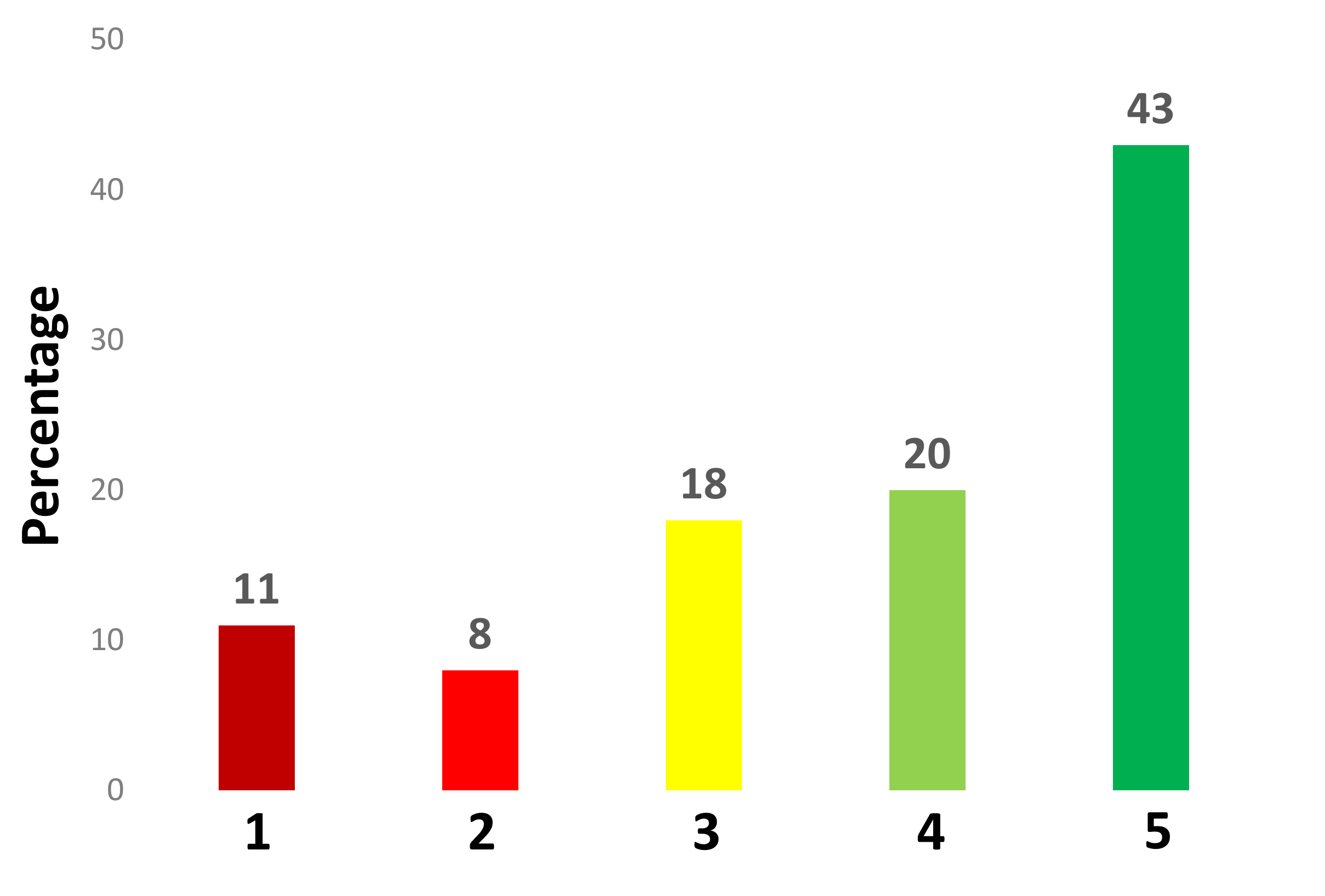}
   \vspace{-4mm}
    \caption{Quality score distribution.}
    \label{fig:scores}
   \vspace{-5mm}
\end{figure}
To test the quality of our crowd-sourced dataset, we perform a limited
user study with 15 testers who are not aware of the nature of the work and how we collect the dataset. We give them a random collection of instances, each of which is a triplet of source graph,  
modification query, and target graph. The tester would then give a score indicating the quality of each instance based on the following two criteria: (i) \textit{how well the 
modification query is reflected in the target graph?} and
(2) \textit{how natural are the query and the graphs?} {Regarding the second criterion, we instruct the scorer to assess whether the query and graph are human-like, grammatically and semantically. Furthermore, as most scorers are knowledgeable in image search, they are also required to evaluate whether they think the query is plausible in a search scenario.}

Figure~\ref{fig:scores} shows the score distribution from 200 randomly chosen instances.
We observe that most of the quality scores of 3 or 4  are due to
the modification query or graphs to be unnatural.  
Testers tend to give the score of 1 for semantically 
wrong instances (e.g the modification query does not match the changes). Overall, the testers
judge the data to be good with the average score of 3.76.

 \subsection{Extension: Multiple Operations}
\label{sec:multiop}
In \S \ref{sec:crowd}, we have introduced our basic modification operations. From the analysis of our in-house search log, more than 95\% of the queries have only one or two nodes, thus a scenario in which more than one edit operation applied is unlikely. Consequently, the instances in Modified MSCOCO and GCC are constructed with one edit operation. However, in some cases, there can be a very long search description
, which leads to the possibility of longer edit operation sequences. This motivates us to create the multi-operation version of a dataset, i.e. the multi-operation graph modification (MGM) task from GCC data. Please refer to the supplementary material for the details of the data creation for MGM.



\section{Methodology}


In this section, we explore different methods to tackle our proposed
problem. By analyzing the results and comparing different models,
we establish baselines and set up the research direction for future work. 
We start by formalizing the problem, and defining the input as well as the expected output along
with the notations. 
We then define our encoder-decoder 
architecture with the focus on our novel modeling characteristics: (i) {the graph encoder with graph-conditioned, sparsely connected transformer} and (ii) {the early and late feature fusion models for combining information from the input text  and
 graph}.


\paragraph{Notations.}
A graph is represented by $\vx^{\mathcal{G}} := (\vx^{\mathcal{N}},\vx^{\mathcal{E}})$. The node set is denoted by $\vx^{\mathcal{N}}:=\{x_1,..,x_{|\vx^{\mathcal{N}}|}\}$ where $|\vx^{\mathcal{N}}|$ is the number of nodes, and $ x_i \in V_{\mathcal{N}}$ where $V_{\mathcal{N}}$ is the node  vocabulary.  
The edge set is denoted by $\vx^{\mathcal{E}}:=\{x_{i,j} | x_i,x_j \in \vx^{\mathcal{N}}, 
x_{i,j} \in V_{\mathcal{E}}\}$
where $V_{\mathcal{E}}$ is the edge  vocabulary.

\subsection{Problem Formulation}

We formulate the task as a conditional generation problem. Formally, given a source graph $\vx^{\mathcal{G}}$ and a modification query $\vy$, one can produce a target graph $\vz^{\mathcal{G}}$ by maximizing
the conditional probability
    $p(\vz^{\mathcal{G}} \mid \vx^{\mathcal{G}},\vy)$.
%
As a graph consists of a list of \emph{typed} nodes and  edges, 
we further decompose the conditional probability \cite{you2018graphrnn} as, 
\begin{equation}
    p(\vz^{\mathcal{G}} \mid \vx^{\mathcal{G}},\vy)=p(\vz^{\mathcal{N}}\mid \vx^{\mathcal{G}},\vy)\times p(\vz^{\mathcal{E}}\mid \vx^{\mathcal{G}},\vy, \vz^{\mathcal{N}}),
\label{eq:cprob1}
\end{equation}
where $\vz^{\mathcal{N}}$ and $\vz^{\mathcal{E}}$ respectively denote the nodes and edges of the graph $\vz^{\mathcal{G}}$.

Given a training dataset of input-output pairs, denoted by $\D \equiv
\{(\vx^{\mathcal{G}}_d, \vy_d, \vz^{\mathcal{G}}_d)\}_{d=1}^D$, we train the model by maximizing the {\em conditional log-likelihood}  
$
\ell_{\text{CLL}} = \ell_{\text{Node}} + \ell_{\text{Edge}}
$
where,
\begin{eqnarray}
\ell_{\text{Node}} &\!\!\! =\!\!\! & \sum_{(\vx,\vy,\vz) \in \D}\log p(\vz^{\mathcal{N}}\mid \vx,\vy;\theta_{\mathcal{N}})\\
\ell_{\text{Edge}} &\!\!\! =\!\!\! &\sum_{(\vx,\vy,\vz) \in \D}\log p(\vz^{\mathcal{E}}\mid \vx,\vy, \vz^{\mathcal{N}}; \theta_{\mathcal{E}}).
\label{equ:obj}
\end{eqnarray}
During learning and decoding, we sort the nodes according to a  topological order which exists for all the directed graphs in our user-generated and synthetic datasets. 


\subsection{Graph-based Encoder-Decoder Model}
\label{sec:enc}

\begin{figure*}[t]
    \centering
    \includegraphics[width=0.81\textwidth]{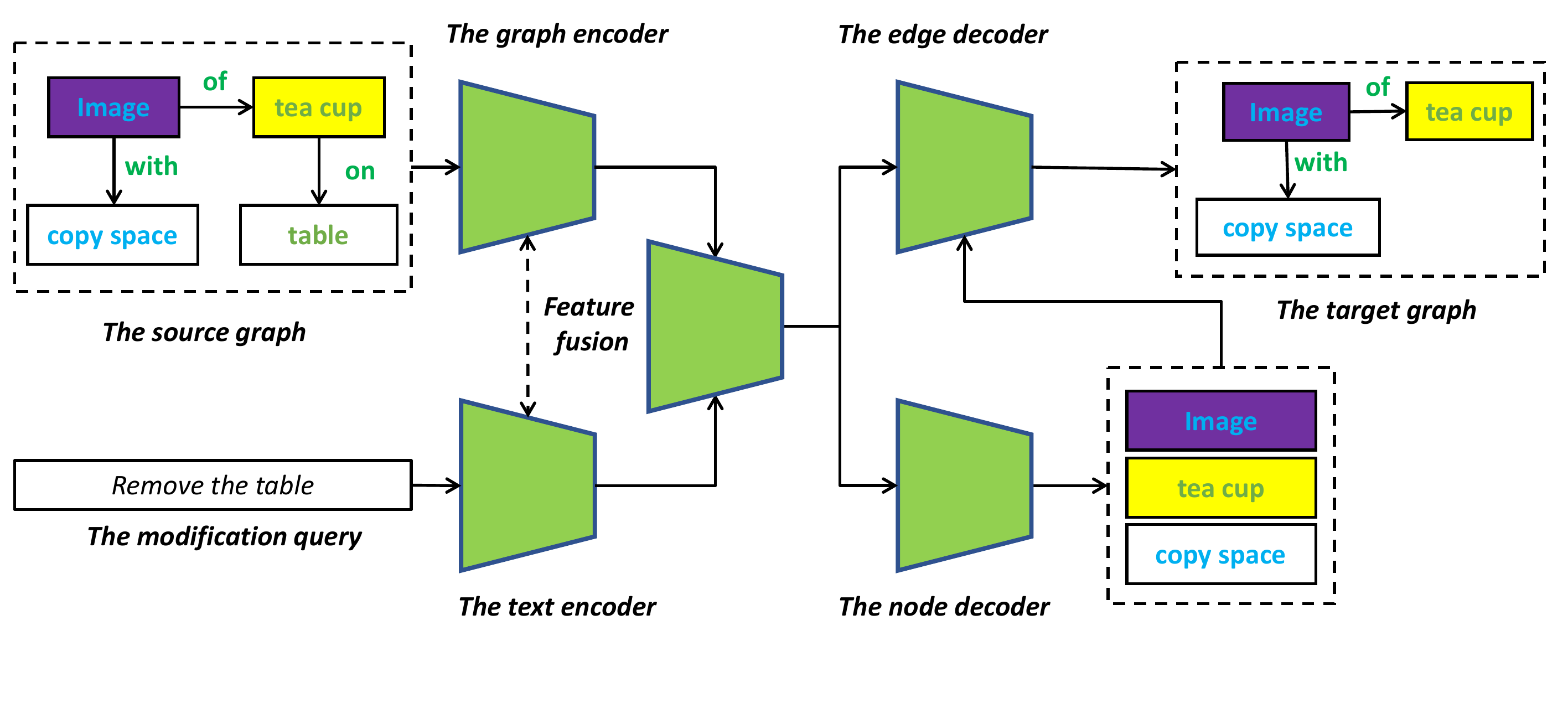}
    \vspace{-8mm}
    \caption{The information flow of our model. Green boxes denote the main
    computational units.}
    \label{fig:architecture}
    \vspace{-5mm}
\end{figure*}

Inspired by the machine translation literature~\cite{bahdanau2014neural,jean2015montreal}, we
build our model based on the encoder-decoder framework.
Since our task takes a source graph and a modification query as inputs, we need two encoders to model the graph and text information separately. Thus,
there are four main components in our model: the query encoder, the graph 
encoder, the edge decoder and the node decoder. The information flow between
the components is shown in \Figref{fig:architecture}. In general,
we encode the graph and text modification query into a joint representation,
then we generate the target graph in two stages. Firstly, the target nodes are 
generated via a node-level recurrent neural network (RNN). Then we leverage 
another RNN to produce the target edges over the nodes.

\subsubsection{Graph Encoder: Sparsely Connected Transformer } 
The standard transformer architecture~\cite{vaswani2017attention,yang2019auto} relies on a grid of fully-connected
self-attention to obtain the contextualized representations from a sequence
of elements. In this work, we propose the graph-conditioned, sparsely
connected transformer to encode the information from a graph.
Our idea is partially inspired by the sparse transformer through 
factorization~\cite{child2019generating}. Despite the similar name, the two
methods share very few similarities in both motivations and mechanisms.
The architecture of our graph encoder with the sparely connected transformer
is detailed below.

Compared to natural language text, graphs are 
structured data, which are comprised of two main components: nodes and edges. 
To efficiently encode a graph, we need to encode the information not only
from these constituent components, but also their interactions, namely the
node-edge association and connectivity. Thus, we incorporate the information from all the edges to the nodes from which 
these edges are originated.
More formally, our edge-aware node embedding $\vx_i$ can be obtained from the list of source graph nodes  and edges via,
%
\begin{equation}
\label{eqn:vx}
    \vx_i = \vT^{\mathcal{N}}[x_{i}] + \sum_{j\in J(i)} \vT^{\mathcal{E}}[x_{ij}],
\vspace{-2mm}
\end{equation}
where $\vT^{\mathcal{N}}$ and $\vT^{\mathcal{E}}$ are the embedding tables for node and edge labels respectively, and $J(i)$ is the set of nodes connected {(both inbound and outbound)} to the $i$th node in the graph.

After getting the edge-aware node embeddings, we employ the sparsely connected transformer to learn the contextualized embeddings of the whole graph.
Unlike the conventional transformer, we do not incorporate the positional encoding into our graph inputs because the nodes are not in a predetermined sequence. Given the edge information from $\vx^{\mathcal{E}}$, we enforce the connectivity information
by making nodes only visible to its \textit{first order neighbor}. Let us denote the attention grid of the transformer as $\textbf{A}$. 
We then define $ A[\vx_i,\vx_j]=f(\vx_i, \vx_j)$ if $x_{i,j} \in \vx^{\mathcal{E}}$ or zero otherwise, 
%
%
%
where $f$ denotes the normalized inner product function. 

The sparsely connected transformer, thus, provides the graph node representations  which are conditioned on the graph structure, using the edge labels in the input embeddings and sparse layers in  self-attention. %
We denote the node representations in the output  of the sparsely connected transformer by $[\vm_{x_1},..,\vm_{x_{|\vx^{\mathcal{N}}|}}]$.


\subsubsection{Query Encoder} 

We use a standard transformer encoder \cite{vaswani2017attention} to encode the modification query $\vy=(y_1,..,y_{|\vy|})$ into  $[\vm_{y_{1}},...,\vm_{y_{|\vy|}}]$.
Crucially, 
in order to encourage semantic alignment, we share the parameters of the graph  and  query encoders.

\subsubsection{ Information Fusion of Encoders}
\label{sec:cross}

In a conventional encoder-decoder model, usually there is only 
one encoder. In our scenario, there are two sources of information, which 
require separate encoders. 
The most straightforward way to incorporate  
the two information sources is through concatenation. 
Concretely, the combined representation would be, \begin{equation}
\label{eqn:combined}
\vm=[\vm_{x_{1}},...,\vm_{x_{|\vx^{\mathcal{V}}|}}, \vm_{y_{1}},...,\vm_{y_{|\vy|}}].
\end{equation}
The decoder component will then be responsible for information  communication between the two encoders through its connections to them. In the following, we propose more advanced methods to combine the two sources of information.

\paragraph{Late Fusion via Gating.} To enhance the ability of the model to combine the encoders' information for a better use of the decoder, we introduce a parametric approach with the gating mechanism. 
Through the gating mechanism, we aim to filter useful information from the graph based on the modification query, and vice versa. 

More specifically, we add a special [CLS] token to the graph and in front of the query sentence. The representation of this token in the encoders  will then  capture the holistic understanding, which we denote by $\vm_{\vx^{\mathcal{G}}}$ and $\vm_{\vy}$ for the graph and modification query respectively.  
%
We make use of these holistic meaning vectors to filter useful information from the representations of the graph nodes $\vm_{x_{i}}$ and    modification query tokens $\vm_{y_{j}}$ as follows,
\begin{eqnarray}
\vg_{x_i} &\!\!\! =\!\!\! & \pmb{\sigma}(\mathrm{MLP}(\vm_{x_{i}}, \vm_{\vy})) \\
\vm'_{x_{i}} &\!\!\! =\!\!\! & \vg_{x_i}\odot \vm_{x_{i}} \\
\vg_{y_j} &\!\!\! =\!\!\! & \pmb{\sigma}(\mathrm{MLP}(\vm_{y_{j}}, \vm_{\vx^{\mathcal{G}}})) \\
\vm'_{y_{j}} &\!\!\! =\!\!\! & \vg_{y_j}\odot \vm_{y_{j}},
\end{eqnarray}
where $\mathrm{MLP}$ is a multi-layer perceptron,  $\odot$ indicates an element-wise multiplication, and $\pmb{\sigma}$ is the element-wise sigmoid function used to construct the gates $\vg_{x_i}$ and $\vg_{y_j}$. The updated node $\vm'_{x_{i}}$ and token $\vm'_{y_{j}}$ are then used in the joint encoders  representation of  \Eqref{eqn:combined}. 

We refer to this gating mechanism as \emph{late fusion} since it does not
let the information from the graph and text interact in their respective lower level encoders. In other words, the fusion  happens after the
contextualized information has already been learned.

\paragraph{Early Fusion via Cross-Attention.} To allow a deeper interaction between the graph and  text encoders, we explore fusing features at the early stage  before the contextualized node $\vm_{x_{i}}$ and token $\vm_{y_{i}}$ representations are learned. This is achieved via \textit{cross-attention}, an early fusion technique.

\begin{figure}[h]
\centering
    \includegraphics[scale=0.3]{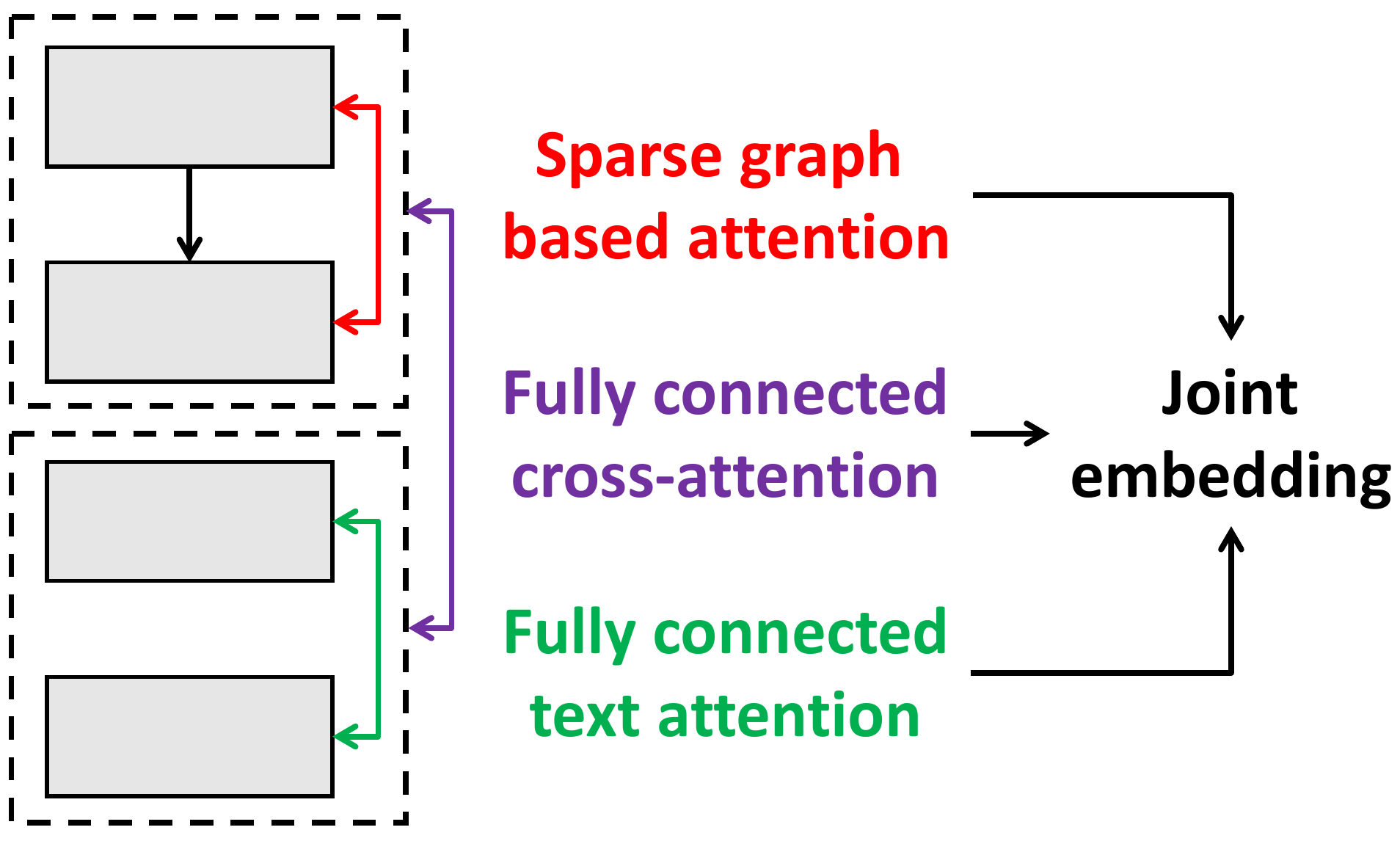}
  \vspace{-2mm}
    \caption{Cross-attention  fusion.}
    \label{fig:cross}
  \vspace{-5mm}
\end{figure}
Recall that the parameters of the graph and query encoders are shared to enable encoding of the two sources in the same semantic space. That is, we use the same transformer encoder for both sources. In cross-attention, we concatenate the $\vx$ (from \Eqref{eqn:vx}) and $\vy$ before rather than after the transformer encoder. 
As such, the encoder's input is $[\vx, \vy]$. 
In the transformer, the representation of each query token gets updated by self-attending to the representations of all the query tokens and graph nodes in the previous layer. 
However, the representation of each graph node gets updated by self-attending only to its graph neighbors according to the connections of the sparsely connected transformer as well as all query tokens.  
%
%
%
The final representation $\vm$ is taken from the  output of 
transformer. \Figref{fig:cross} shows the information flow in 
the cross-attention mechanism.




\subsubsection{Node-level Decoder} We use GRU cells \cite{cho2014learning} for our RNN decoders. The node-level decoder is a vanilla auto-regressive model described as, 
\begin{align}
&\vh^{\mathcal{N}}_{t} = \mathrm{GRU}^{\mathcal{N}}(z_{t-1}, \vh^{\mathcal{N}}_{t-1})\\
\label{eq:nodeRNN1}
&\vc^{\mathcal{N}}_{t} = \mathrm{ATTN}^{\mathcal{N}}(\vh^{\mathcal{N}}_{t}, \vm) \\
\label{eq:nodeRNN2}
&p(z_t\mid \vz_{< t}, \vx^{\mathcal{G}},\vy) = \\
&\quad\quad\quad\quad\quad\softmax(\vW[\vh^{\mathcal{N}}_t,\vc^{\mathcal{N}}_t]+\vb),
\label{eq:nodeRNN3}
\end{align}
where $\vz_{< t}$ denotes the nodes generated before  time step $t$,  $\mathrm{ATTN}^{\mathcal{N}}$ is a Luong-style attention \cite{luong2015effective}, and $\vm$ is the memory vectors  from information fusion of the encoders (see \S \ref{sec:cross}).


\subsubsection{Edge-level Decoder} 

\begin{figure}[h]
\centering
    \includegraphics[scale=0.45]{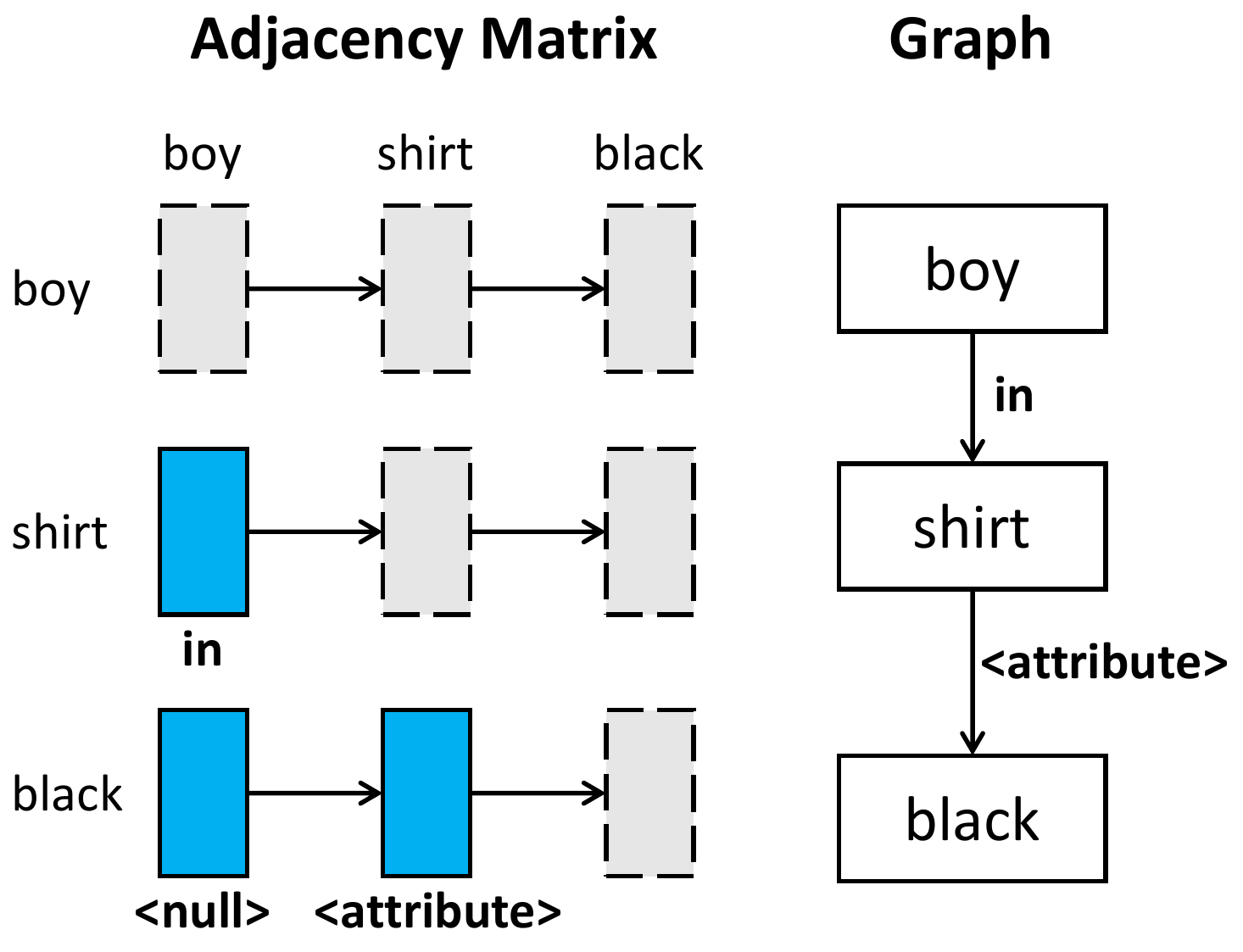}
   \vspace{-3mm}
    \caption{Adjacency matrix style decoder.}
    \label{fig:adj}
   \vspace{-3mm}
\end{figure}

For the edge decoder, we  first use an adjacency-style generation  \cite{you2018graphrnn}.
The rows/columns of the adjacency matrix are labeled by the nodes in the order that they have been generated by the node-level decoder. 
For each row, we have an auto-regressive decoder which emits the label of each edge to other nodes from the edge vocabulary, including a special token [NULL] showing an edge does not exist. 
As shown in \Figref{fig:adj}, 
we are only interested in the lower-triangle part of the matrix, as we assume that the node decoder has generated the nodes in a topologically sorted manner. 
The dashed upper-triangle part of the adjacency matrix are used only for parallel computation, and they will be discarded. 

We use an attentional decoder using GRU units for generating edges. It operates similarly to the node-level decoder using \Eqref{eq:nodeRNN1} and \Eqref{eq:nodeRNN2}.
For more accurate typed edge generation, however, we incorporate the hidden states of the source and target nodes (from the node decoder) as inputs when updating the hidden state of the edge decoder:
\begin{eqnarray}
\vh^{\mathcal{E}}_{i,j} &\!\!\! =\!\!\! & \mathrm{GRU}^{\mathcal{E}}(z_{i,j-1},\vh^{\mathcal{N}}_{i},\vh^{\mathcal{N}}_{j}, \vh^{\mathcal{E}}_{i,j-1}),
\end{eqnarray}
where $\vh^{\mathcal{E}}_{i,j}$ is the hidden state of the edge decoder for row $i$ and column $j$, and $z_{i,j-1}$ is the label of the previously generated edge  from node $i$ to $j-1$.


%





\begin{figure}[h]
\centering
   \vspace{-4mm}
     \includegraphics[scale=0.4]{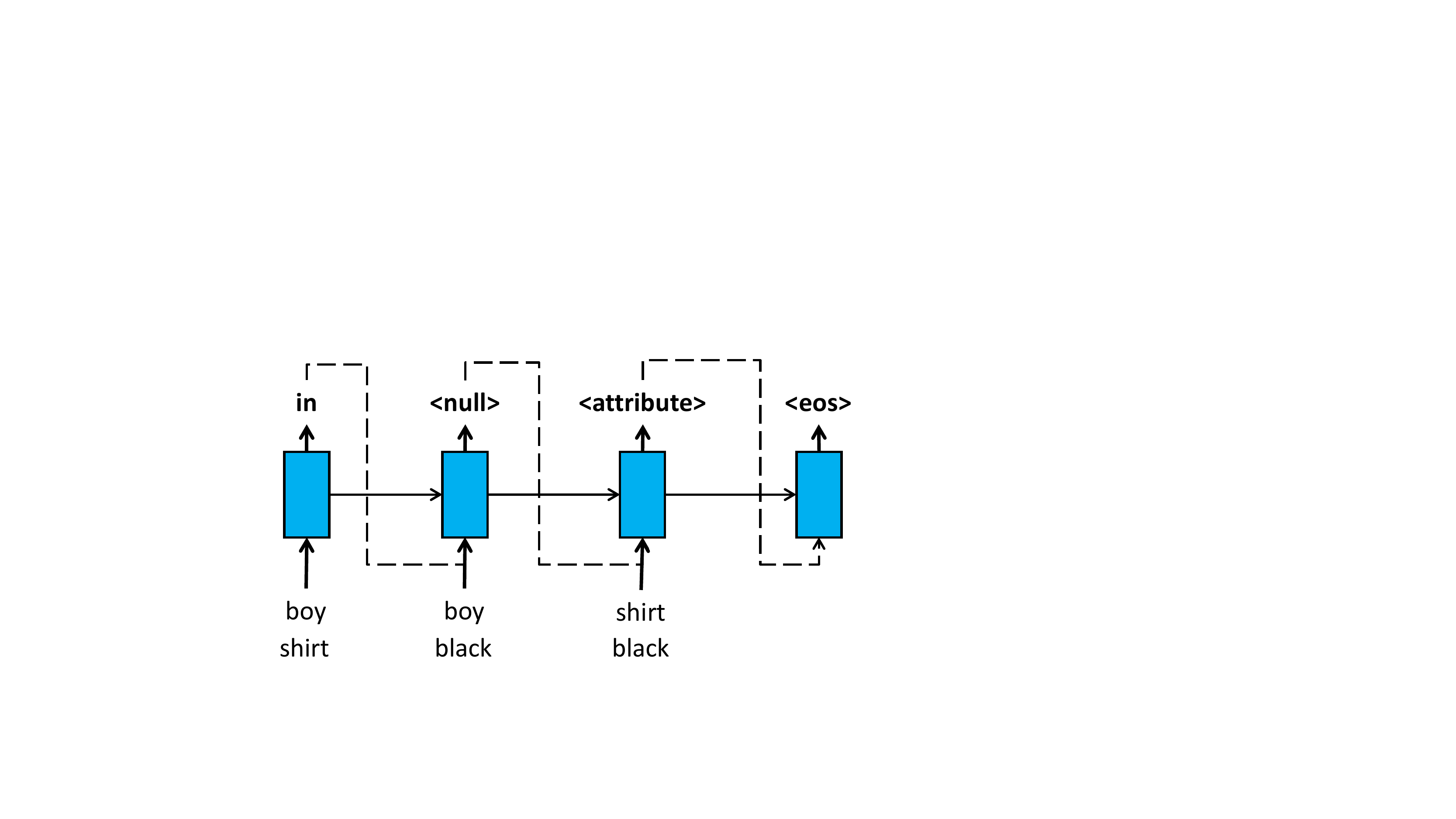}
   \vspace{-1mm}
    \caption{A flat edge-level decoder. } 
    \label{fig:flat}
   \vspace{-3mm}
\end{figure}
However, there are two drawbacks in this edge generation method. Firstly, the dummy edges in the adjacency matrix cause a waste of  computation.
Secondly, the edges generated by the previous rows are not conditioned upon when the edges in the next row are generated.  
However, it may be beneficial to use the information  about the outgoing edges of the previous nodes to enhance the generation accuracy of  the outgoing edges of the next node. We will analyze this hypothesis in \S 4. 
%
%
Hence, we suggest flattening the lower-triangle of the adjacency matrix. 
We remove the dummy edges and concatenate the rows of the lower triangular matrix to form a sequence of pairs of nodes for which we need to generate edges (Figure \ref{fig:flat}).
This strategy results in  using information about \emph{all} previously generated edges when a new edge is generated.

\begin{table*}[t]
\small{
    \centering
    \begin{tabular}{l||ccc||ccc}
    \toprule
    & \multicolumn{3}{c||}{Synthetic Data} & \multicolumn{3}{c}{User-Generated Data} \\
    & Edge F1 & Node F1 & Graph Acc & Edge F1 & Node F1 & Graph Acc
   \\
              \midrule\midrule
    \textbf{Baselines} & & & & & &\\
      Copy Source &64.62 & 78.41 & -  &31.42 & 66.17 & -\\
      Text2Text & 72.74 & 91.47 &   64.42 & 52.68 & 78.59 & 52.15 \\
      Modified GraphRNN~\cite{you2018graphrnn} & 55.76 & 80.64 & 50.72 & 57.17 & 80.68 & 56.75 \\
      Graph Transformer~\cite{cai2019graph} & 75.68 & 91.21 & 71.38 & 59.43 & 81.47 & 58.23 \\
      DCGCN~\cite{guo2019densely} & 72.47 & 89.08 & 68.89 & 54.23 & 79.05 & 52.67 \\
     \midrule
    \textbf{Our Models} & & & & & &\\
      Fully Conn Trans + Adj Matrix + Concat & \;\;76.49* & 91.54  & \;\;72.13* & 57.47 & 81.29 & 56.91 \\
      Sparse Trans  + Adj Matrix  + Concat& \;\;77.94* & 91.94  & \;\;74.68* & 57.78 & 81.36  & 56.98\\
      Sparse Trans + Flat-Edge  + Concat & \;\;79.13* & 92.11 &\;\;76.13* & 57.92 & 81.74 &57.03\\
      Sparse Trans + Flat-Edge + Gating  &\;\;80.13*  & \;\;92.54*  &  \;\;77.04* &\;\;59.58*  & \;\;82.39*  &  \;\;59.63*\\
      Sparse Trans + Flat-Edge + Cross-Attn &  \;\;\textbf{86.52}* & \;\;\textbf{95.40}*&   \;\;\textbf{82.97}*& \;\;\textbf{62.10}* & \;\;\textbf{83.69}*&  \;\;\textbf{60.90}*\\
     \bottomrule
    \end{tabular}
    \caption{Node-level, edge-level and graph-level matching score (\%) over two datasets (modified from MSCOCO).  ``*'' indicates statistically significant difference (p \textless 0.0001) from the best baseline.}
    \label{tab:synthdata}
    }
\end{table*}

\section{Experiments}
\paragraph{Baselines.} We consider five baselines for comparison. In 
 ``Copy Source'' baseline (i), the system copies the source graph to the target graph\footnote{
 It is based on the observation that the user only modifies a small portion of the source graph.}.
 In the ``Text2Text'' baseline (ii), we flatten the graph and {reconstruct the natural sentence} similarly to the modification query. In the
``Modified  GraphRNN'' baseline (iii),
we use the breadth-first-search (BFS) based node ordering to flatten the graph\footnote{The topological ties are broken by the order of the nodes appearing in the original query.}, and use
RNNs as the encoders~\cite{you2018graphrnn} and a decoder similar to our systems. In the final two baselines,
``Graph Transformer'' (iv) and ``Deep Convolutional Graph Networks'' (DCGCN) (v), we use the Graph Transformers~\cite{cai2019graph} and Deep Convolutional Graph Networks~\cite{guo2019densely} to encode the source graph  (the decoder is identical to ours).

\paragraph{Our Model Configurations.} We report the results of different  configurations of our model. 
The ``Fully Connected Transformer'' uses dense connections for the graph encoder. This is in contrast to ``Sparse Transformer'', which uses the connectivity structure of the source graph in self attention  (see \S 3.2.1).  
%
%
%
The information from the graph and query encoders can be combined by ``Concatenation'', late fused by ``Gating'', or early fused by ``Cross Attention'' (see \S 3.2.3). 
The ``Adjacency Matrix'' style for edge decoding can be replaced with ``Flat-Edge'' generation (see \S 3.2.5). 

\paragraph{Evaluation Metrics.}
We use two automatic metrics for the evaluation. Firstly, we calculate the
precision/recall/F1-score of the generated nodes and edges. 
Secondly, we use 
the strict-match accuracy, which requires 
the generated graph to be identical to the target graph for a correct prediction.

\paragraph{Data Splits.}
We partition the synthetic MSCOCO data into 196K/2K/2K for training/dev/test, and GCC data into 400K/7K/7K for training/dev/test.  
%
We randomly split the crowdsourced user-generated data into 30K/1K/1K for training/dev/test. 

\subsection{Experimental Results}
\label{sec:syn_expr}
 



Table \ref{tab:synthdata} reports the results of our model and the baselines on the synthetic and user-generated datasets.
From the experimental results, various configurations of our model are superior to the baselines by a significant margin. Noticeably, DCGCN and graph transformer are strong baselines, delivering SOTA performance across tasks such as AMR-to-text generation and syntax-based
neural machine translation \cite{guo2019densely, cai2019graph}. 
We believe the larger number of edge types in our task impairs their capability. 

We ablate the different components of the proposed methods to appraise their effectiveness (\cf the bottom pane of \tabref{tab:synthdata}). First, our hypothesis about the preference of flat-edge generation over adjacency matrix-style edge generation is confirmed. 
Furthermore, the two-way communication between the graph and query encoders through the gating mechanism consistently outperforms a simple concatenation in terms of both edge-level and node-level generation. 
Eventually the cross-attention -- the early fusion mechanism, leads to substantial improvement in all metrics.

We also observe that generating the graphs for the crowdsourced data is much harder than the synthetic data, which we believe is caused by  diversity in semantics and expressions introduced by the annotators. Consequently, all models suffer from performance degradation.
 %
\setlength\tabcolsep{1.5pt}
\begin{table}[t]
\vspace{-3mm}
\small{
    \centering
    \begin{tabular}{l||ccc||ccc}
    \toprule
    & \multicolumn{3}{c||}{MSCOCO} & \multicolumn{3}{c}{GCC} \\
    & {\tiny Edge F1} & {\tiny Node F1} & {\tiny Graph Acc} & {\tiny Edge F1} & {\tiny Node F1} & {\tiny Graph Acc}
   \\
              \midrule\midrule
        Graph Trans. & 75.68 & 91.21 & 71.38 & 42.76 & 82.38 & 34.31\\      
      Concat & 79.13 & 92.11 & 76.13 & 45.09 & 86.93 &37.53\\
       Gating  &80.13  & 92.54  & 77.04 & 52.85  & 91.60  & 45.79\\
      Cross-Attn & 86.52 & 95.40&  82.97& \textbf{57.68} & \textbf{93.84} & \textbf{52.50}\\
     \bottomrule
    \end{tabular}
    \caption{Node/Edge/Graph level matching scores comparing the best baseline - Graph Transformer to our model variants on synthetic MSCOCO and GCC. }
    \label{tab:gccdata}
    \vspace{-3mm}
    }
\end{table}

\newcolumntype{Y}{>{\centering\arraybackslash}X}


\newcommand*\rot[1]{\rotatebox{90}{#1}}

\begin{table}[t]
 \small{
    \centering
    \setlength\tabcolsep{1.5pt} 
    \begin{tabular}{l||c|c|c||c|c|c}
    \toprule
    & \multicolumn{3}{c||}{MOPs (avg. 1.44)} & \multicolumn{3}{c}{MOPs (avg. 2.01)}\\
    & 1-2 & 3-4 &5+&1-2 & 3-4 &5+
   \\
              \midrule\midrule

    Text2Text &27.40&0.87&0.00& 26.84&2.38&0.24\\
    M. GraphRNN & 26.10&0.64&0.00&25.17&1.81&0.00\\
    Graph Trans. & 29.97&1.75&0.00&29.14&4.26&0.53\\
    Cross-Attn &  \textbf{47.95}&\textbf{14.82}&\textbf{1.01}&\textbf{49.93}&\textbf{19.77}&\textbf{4.45}\\
     \bottomrule
    \end{tabular}
    \caption{Graph-level accuracy on two multiple operations (MOPs) datasets, one with an average of 1.44 operations per query, the other with 2.01}
    \label{tab:multiop}
    \vspace{-3mm}
     }
\end{table}

Nevertheless, the performance trends of different configurations of our model
are almost identical on the user-generated and synthetic data. 
%
{Finally, \Tabref{tab:gccdata} indicates with the increase of the complexity of graphs, the models have a difficulty in inferring the relations among nodes for GCC data, which causes a dramatic drop in terms of the edge F1 score and graph accuracy.}

\subsection{Multi-Operation Performance}
To study the multiple operations scenario, 
we create two datasets\footnote{Please refer to Appendix~\ref{sec:mulop} for the data creation process.} where the average number of the operations are 1.44 and 2.01. For each dataset, we train the baselines and our methods on the full training set. The test set is grouped into four bins according to the number of operations.

According to \Tabref{tab:multiop}, all models demonstrate sharp decreases in performance with the increase of the number of operations. Our model still performs significantly better than the baselines. Having said that, for more than 1-2 operations, all models do not perform satisfactorily, prompting further research.


\begin{figure*}
    \centering
    \includegraphics[width=0.8\textwidth]{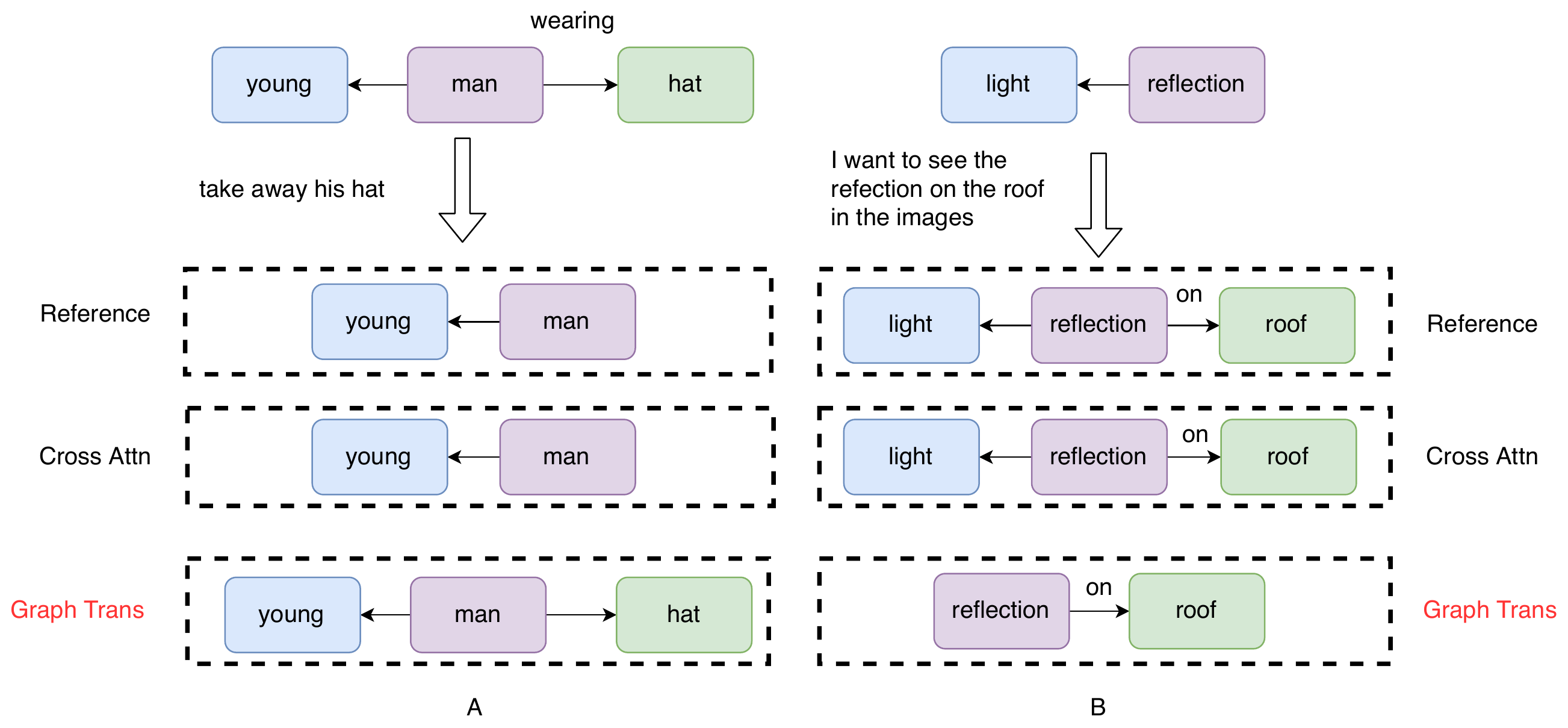}
    \caption{Our best model vs the Graph Transformer.}
    \label{fig:quali_analysis}
\end{figure*}


\subsection{Quantitative Analysis}

The best configuration of our model is based on cross-attention, with flat-edge decoder, and sparse transformer. 
We investigate which cases this configuration outperforms the baselines. 
{As seen in Figure \ref{fig:quali_analysis}, cross-attention is able to understand the \textit{pronoun} and correctly removes the connected object and its associated relation as evidenced by the first example \textbf{A}. In addition, example \textbf{B} demonstrates when graph transformer observes a longer description, it lacks the capability of fusing the semantics between the source graph and the modification query; then certain nodes from the source graph are not preserved. We believe that the proposed approach can reduce the noise in graph generation, and retain fine-grained details better than the baselines.}

\section{Related Work}
Semantic parsing is a sequence-to-graph transduction task, mapping natural language sentences to their meaning representation, e.g. see \cite{buys-blunsom-2017-robust,iyer-etal-2017-learning,dong-lapata-2018-coarse}; this is different from our graph conditional semantic parsing. 
Recently, context-dependent semantic parsing has gained attraction \cite{iyyer2017search, srivastava2017parsing, suhr2018learning, he2019pointer}. Our work focuses on the update of scene graphs based on users' queries, while previous works model the modifications of semantic representations in multi-turn dialogue systems.
Due to their effectiveness, GCNs and graph transformer have been used as graph encoder for graph-to-sequence transduction in semantic-based text generation \cite{bastings-etal-2017-graph,beck2018graph, guo2019densely, cai2019graph,song2018graph,wu2020comprehensive}.

\section{Conclusion}
In this paper, we explore a novel problem of conditional graph 
modification, in which a system needs to modify a source graph
according to a modification command. Our best system,
which is based on graph-conditioned transformers and cross-attention
information fusion, outperforms strong baselines adapted from machine
translations and graph generations. The code and datasets will be released to encourage further research in this direction.

\section*{Acknowledgments}
We would like to thank anonymous reviewers and all voluntary scorers for their valuable feedback and suggestions. The computational resources of this work are supported by Adobe and the Multi-modal Australian ScienceS Imaging and Visualisation Environment (MASSIVE) (\url{www.massive.org.au}). This work is partially supported by an ARC Future Fellowship (FT190100039) to G. H.

\bibliography{emnlp2020}
\bibliographystyle{acl_natbib}

\appendix

\section{Training Details}
Our encoder is comprised of 3 stacked sparse transformers, with 4 heads at each layer. The embedding size is 256, and the inner-layer of feed-forward networks has a dimension of 512. Both node-level and edge-level decoders are one-layer GRU-RNN with a hidden size of 256, and the size of embeddings are 256 as well. We train 30 epochs and 300 epochs for synthetic and user-generated data respectively, with batch size of 256. We evaluate the model over the dev set every epoch, and choose the checkpoint with the best graph accuracy for the inference. We run all experiments on a single Nvidia Tesla V100.

\Tabref{tab:time} shows that our cross-attention model is more efficient than other models in terms of GPU computing time. \Tabref{tab:params} displays the number of parameters used for each model.

%
\begin{table}[h]
\small{
    \centering
    \begin{tabular}{l||cc||c}
    \toprule
    & \multicolumn{2}{c||}{MSCOCO} & GCC \\
    & syn & crowd. & syn.\\
              \midrule\midrule
       DGCN & 205& 325& 712\\
        Graph Trans. & 126& 114& 478\\      
      Concat & 146 & 160& 457\\
       Gating  & 248 &248 & 534\\
      Cross-Attn & ~~93 & 132 & 469\\
     \bottomrule
    \end{tabular}
    \caption{GPU time (ms/step) over different settings at training stage.}
    \label{tab:time}
    }
\end{table}

\begin{table}[h]
\small{
    \centering
    \begin{tabular}{l||cc||c}
    \toprule
    & \multicolumn{2}{c||}{MSCOCO} & GCC \\
    & syn & crowd. & syn.\\
              \midrule\midrule
       DGCN & 10.9M& 16.3M & 19.3M\\
        Graph Trans. & ~~9.9M & 15.2M& 18.4M\\      
      Concat & ~~8.1M & 12.8M& 15.4M\\
       Gating  & ~~8.1M &12.8M & 15.4M\\
      Cross-Attn & ~~8.1M & 12.8M & 15.4M\\
     \bottomrule
    \end{tabular}
    \caption{Number of parameters over different settings.}
    \label{tab:params}
    }
\end{table}

\section{Performance on validation set}

Table~\ref{tab:dev} shows the performance on the validation/dev set of our models and the best baseline. In general, there is no significant difference between the performance trend in the dev set and the test set.

\begin{table*}[h]
    \centering
  
    \begin{tabular}{l|cc|cc|cc}
       \toprule
    & \multicolumn{4}{c|}{Synthetic} &\multicolumn{2}{c}{\multirow{2}{*}{Crowsourced}}\\
    \cline{2-5}\cline{2-5}
    &\multicolumn{2}{c|}{MSCOCO} & \multicolumn{2}{c|}{GCC}& & \\
    & dev& test & dev& test  & dev& test\\
    \midrule
    Graph Trans. &72.01	&71.38	&40.14	&34.31	&59.88	&58.23\\
    Concat &77.34	&76.13	&43.16	&37.53	&59.20	&57.03\\
    Gating & 78.06	&77.04	&52.87	&45.79	&60.48	&59.63\\
    Cross-Attn. &\textbf{84.12}	&\textbf{82.97}	&\textbf{60.21}	&\textbf{52.50}	&\textbf{61.20}	&\textbf{60.90}\\
    
    \bottomrule
    \end{tabular}
    \caption{Results of the best baseline and our models on dev and test splits. }
    \label{tab:dev}
\end{table*}     

\section{Data Statistics}
\label{sec:data_stat}
\begin{figure}[H]
    \centering
    \includegraphics[scale=0.5]{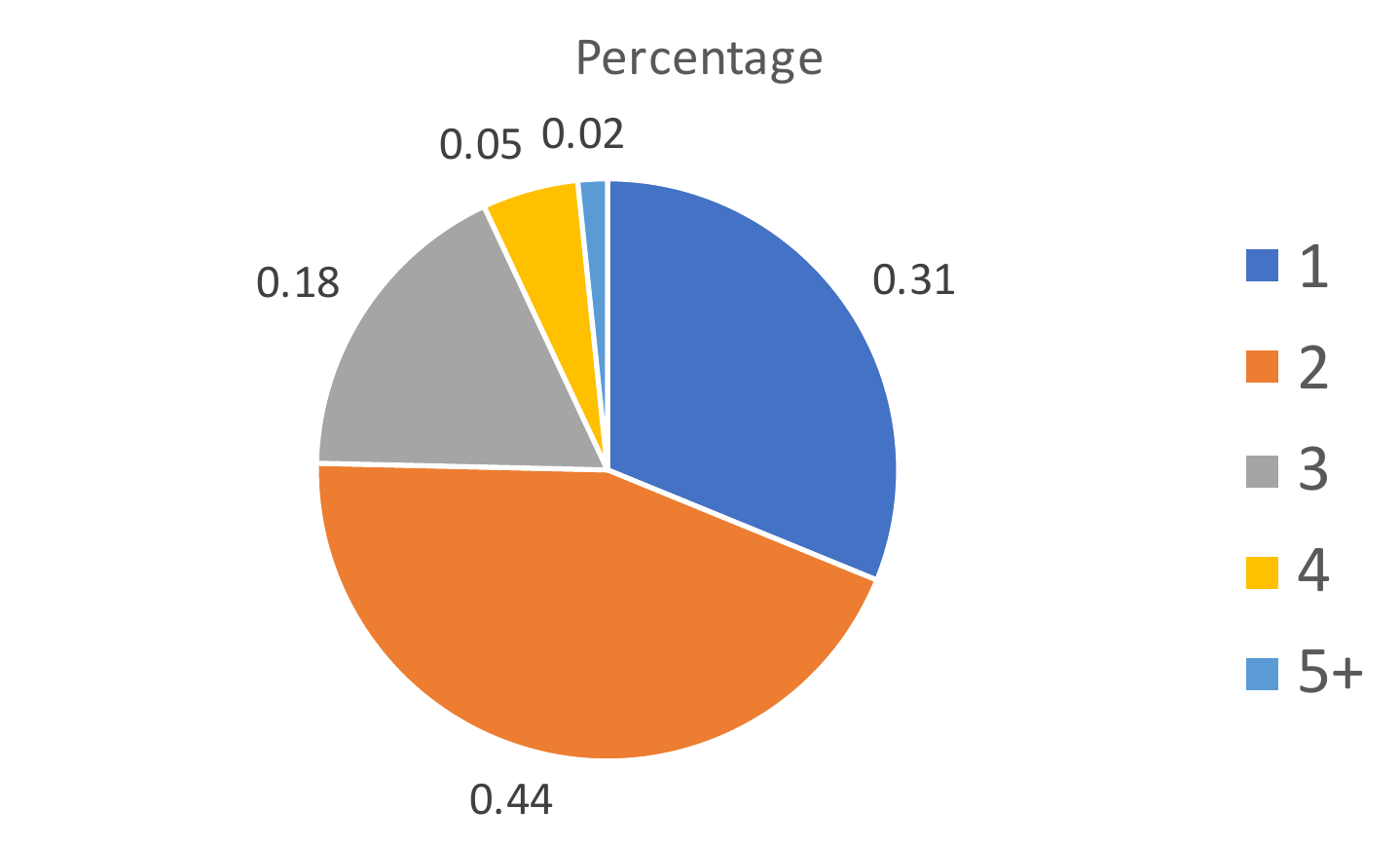}
    \caption{The percentage of the length of queries from in-house search log.}
    \label{fig:q_len}
\end{figure}

As shown in \Tabref{tab:data-stat}, the graph size distributions of source graphs and target graphs are almost identical among the sets. With the increase in text description length, the source graphs become more complicated accordingly. According to \Figref{fig:q_len}, the length of search queries are likely to be less than 5 tokens. Thus,
in a real application, it is unlikely to encounter large graphs (\textgreater3 
nodes) and long modification queries.

\begin{figure}[h!]
     \centering
     \begin{subfigure}[b]{0.4\textwidth}
         \centering
         \includegraphics[width=\textwidth]{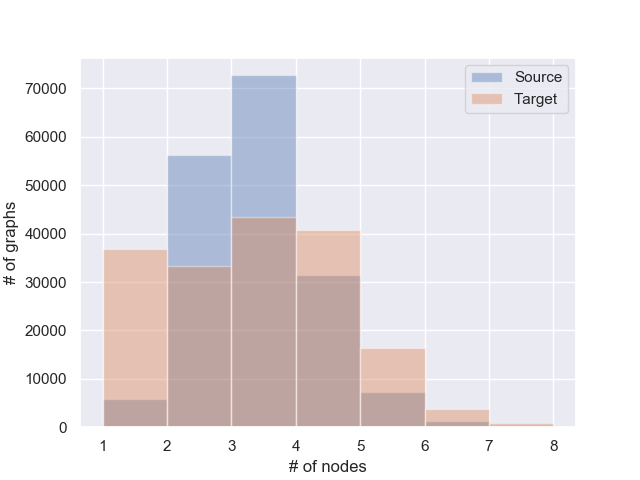}
         \caption{Distribution of nodes on synthetic data}
         \label{fig:node_syn}
     \end{subfigure}
     \begin{subfigure}[b]{0.4\textwidth}
         \centering
         \includegraphics[width=\textwidth]{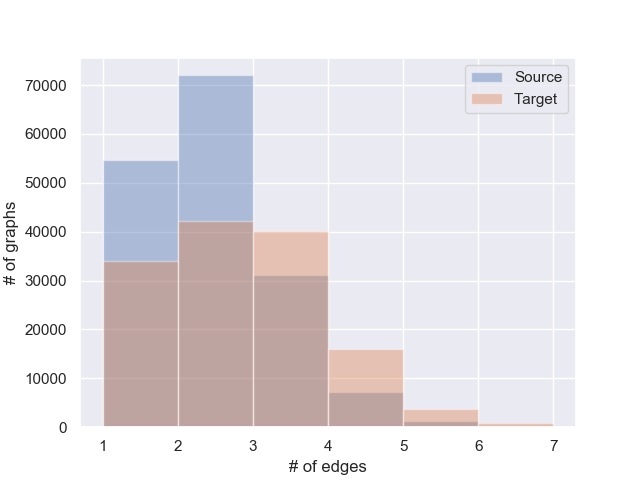}
         \caption{Distribution of edges on synthetic data}
         \label{fig:edge_syn}
     \end{subfigure}
     \begin{subfigure}[b]{0.4\textwidth}
         \centering
         \includegraphics[width=\textwidth]{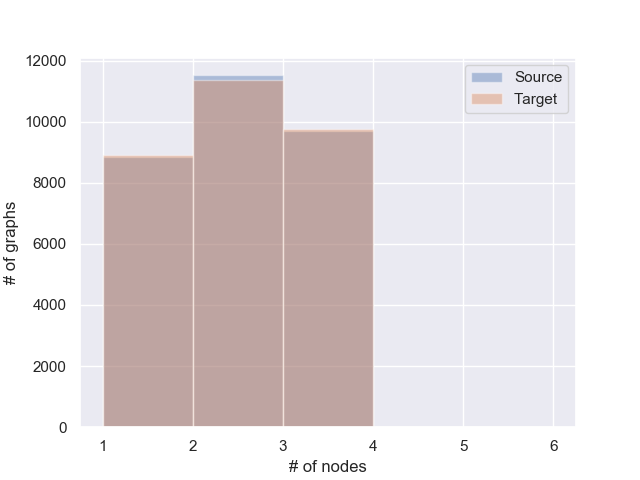}
         \caption{Distribution of nodes on user-generated data}
         \label{fig:node_real}
     \end{subfigure}
     \begin{subfigure}[b]{0.4\textwidth}
         \centering
         \includegraphics[width=\textwidth]{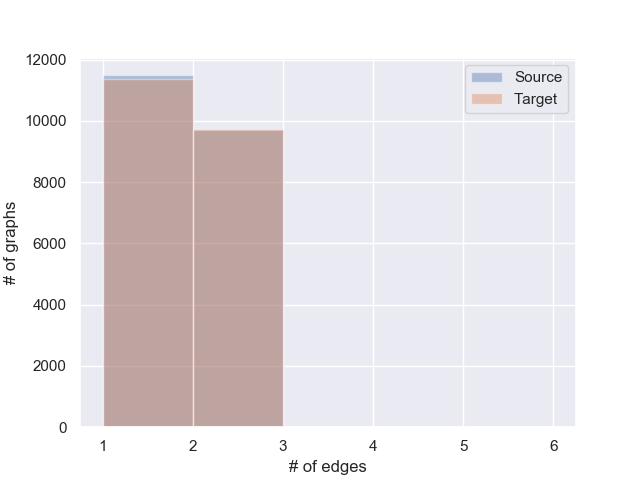}
         \caption{Distribution of edges on user-generated  data}
         \label{fig:edge_real}
     \end{subfigure}
        \caption{The distributions of the number of nodes and edges on synthetic and user-generated data among source graphs and target graphs.}
        \label{fig:dist_node_edge}
\end{figure}

We plot the distributions of the number of nodes and edges on synthetic and user-generated data in \Figref{fig:dist_node_edge}. 
\begin{table*}[h]
    \centering
    \small{
    \begin{tabular}{l|c|c|c}
       \toprule
    & \multicolumn{2}{c|}{Synthetic (Train/Dev/Test)} & Crowsourced \\
    \cline{2-3}\cline{2-3}
    & MSCOCO & GCC & Train/Dev/Test \\
    \midrule
    $\;$ size & 196k~/~2k~/~2k  & 400k~/~7k~/~7k &30k~/~1k~/~1k\\\hline
    $\;$  Ave. \#tokens / text desc & 5.2~/~5.2~/~5.2 & 10.1~/~10.1~/~10.2 &4.8~/~4.8~/~4.8\\\hline
    $\;$  Ave. \#nodes / src graph & 2.9~/~2.9~/~2.9& 3.8~/~3.8~/~3.8&2.0~/~2.0~/~2.0\\\hline
    $\;$  Ave. \#edges / src graph  & 1.9~/~1.9~/~1.9 &2.9~/~2.8~/~2.8&1.0~/~1.0~/~1.0\\\hline
  $\;$  Ave. \#nodes / tgt graph & 2.9~/~2.9~/~2.8&3.8~/~3.8~/~3.8 & 2.0~/~2.0~/~2.0\\\hline
    $\;$  Ave. \#edges / tgt graph & 1.9~/~1.9~/~1.8&2.9~/~2.8~/~2.8 &1.0~/~1.0~/~1.0 \\\hline
    $\;$  Ave. \#tokens / src query & 4.7~/~4.8~/~4.7&4.9~/~4.8~/~4.9&10.1~/~10.2~/~10.0\\
    \bottomrule
    \end{tabular}
    }
    \caption{Statistics of the created datasets. }
    \label{tab:data-stat}
    \vspace{-3mm}
\end{table*}     

\section{Data Creation for Multi-operation Graph Modification}
\label{sec:mulop}
We develop a procedure to create data for the multi-operation graph modification (MGM) task. First of all, we assume that MGM requires at least one operation on the source graph. Then we use four actions (\textit{terminate}, \textit{ins}, \textit{del}, \textit{sub}) paired with a heuristic algorithm to further perform operations on the modified graph. We sample an action, and execute it on the last modified graph until the \textit{terminate} is sampled or we exhaust the available nodes. Intuitively, a large graph can support more modifications, while a smaller graph does not have too much freedom. In addition, we also assume that the modified nodes should not be changed again. 
Hence, the probability of \textit{terminate} should be increased as the edit sequence gets longer, whereas the probabilities of other actions should drop. The heuristic algorithm is summarized in \Algref{alg:multi}. It is worth noting that \Algref{alg:multi} gives us a dataset with different edit sequence lengths.

\begin{algorithm} 
\caption{Multiple operations for graph modifications} 
\label{alg:multi} 
\begin{algorithmic}[1] 
\REQUIRE $G$: scene graphs, $I$: insertion templates, $D$: deletion templates, $S$: substitution templates
\ENSURE $X$: source graphs, $Z$: target graphs, $Y$: modified queries, 
        \STATE{$X$ $\leftarrow$ $\{\}$, $Z$ $\leftarrow$ $\{\}$, $Y$ $\leftarrow$ $\{\}$}
        \STATE{$A$ $\leftarrow$ $\{ins, del, sub\}$,}
         \FOR{$k=1$ \TO $|G|$}
         \STATE{$g$ $\leftarrow$ $G_k$}
         \STATE{$a$ $\sim$ uniform($A$)}
         \IF {$a$ == $ins$} 
            \STATE{$s, q, t$ $\leftarrow$ $\mathrm{insertion}(g, I)$}
            \label{line:ins}
        \ELSIF {$a$ == $del$}
            \STATE{$s, q, t$ $\leftarrow$ $\mathrm{deletion}(g, D)$}
             \label{line:del}
        \ELSE
          \STATE{$s, q, t$ $\leftarrow$ $\mathrm{substitution}(g, S)$}
          \label{line:sub}
        \ENDIF
        
        \STATE{$A$ $\leftarrow$ $\{terminate, ins, del, sub\}$,}
        \STATE{$w$ $\leftarrow$ $\{P, 1, 1, 1\}$,} \COMMENT{$P$ controls the average number of operations.}
        \WHILE{True}
        \STATE{total $\leftarrow$ $\mathrm{TotalNode(t)}$}
        \STATE{avail $\leftarrow$ $\mathrm{AvailableNode(t)}$}
         \IF {len(avail) == 0}
            \STATE{break}
        \ENDIF
        \STATE{$D$ $\leftarrow$ $softmax(w^{\frac{total-avail}{total*\tau}})$}
        \STATE{a $\sim$ $sample$($A$, $D$)}
        \IF {$a$ == $terminate$}
            \STATE{break}
        \ELSIF {$a$ == $ins$} 
            \STATE{$s, q', t$ $\leftarrow$ $\mathrm{insertion}(s, t, I)$}
        \ELSIF {$a$ == $del$}
            \STATE{$s, q', t$ $\leftarrow$ $\mathrm{deletion}(s, t, D)$}
        \ELSE
          \STATE{$s, q', t$ $\leftarrow$ $\mathrm{substitution}(s, t, S)$}
        \ENDIF
        \ENDWHILE
        \STATE {$q$ $\leftarrow$ $\mathrm{concat(q,q')}$}
         \STATE{$X$ $\leftarrow$ $X\cup \{s\}$, $Z$ $\leftarrow$ $Z\cup \{t\}$, $Y$ $\leftarrow$ $Y\cup \{q\}$}
        \ENDFOR
         \STATE{return $X$, $Z$, $Y$}
\end{algorithmic}
\end{algorithm}

\section{Mixing Synthetic and User-generated Data}

Getting annotation from users is expensive, especially for a complex task like
our graph modification problem. Thus, we explore the possibility of \emph{augmenting}
the  user-generated data with synthetic data in order to train a better model. 
However,
one needs to be careful with data augmentation using synthetic data as it inevitably has a different distribution. 
This is evident when we
test the model trained using the synthetic data on the user-generated data. 
The graph generation
accuracy drops to  around 20\%, and adding more synthetic data does not help. 
To efficiently mix the data distributions, we up-sample the user-generated data and mix it with synthetic data with a ratio of 1:1 in each mini-batch. 
%
%

We compare data augmentation using upsampling with transfer learning --  another method to learn from both synthetic and user-generated data~\cite{openai2019solving}. We pretrain our model using the synthetic data, and then fine-tune it on the user-generated data.


\setlength\tabcolsep{1.5pt}
\begin{table}[h]
\small{
    \centering
    \begin{tabular}{l||c|c|c|c|c}
    
              \toprule
      Synthetic Data Size & 30k & 60k&90k& 120k & 150k\\
              \midrule
    Trained with: & & & & & \\          
     \ \ \ \ Synthetic only  &21.63 & 19.33 & 22.03 & 22.40&18.87\\
     \ \ \ \ Pretrain-Finetune & 61.40 & 63.37  & 63.87 & 62.300& 63.47\\
     \ \ \ \ Data Augment. & 70.27 & 72.37 &74.80 & 74.67 &75.23\\
     \bottomrule
    \end{tabular}
    \caption{Graph accuracy (\%) over different data settings. 30k means adding 30k synthetic instances.}
    \label{tab:mix}
}
\end{table}
Table  \ref{tab:mix} reports the results. It shows that data augmentation with up-sampling is a very effective method to leverage both sources of data, compared to transfer learning. Also, as the size of the synthetic data increases, our proposed scheme further  improves the performance to a certain point where it plateaus. More specifically, the performance reaches  plateau after injecting 90k instances (the data ratio of 3:1). 
Both up-sampling and pre-training lead to better models compared to 
using only synthetic or user-generated data.
The graph accuracy for model trained only on user-generated data is 60.90\% (see the best result from Table 1 in the main paper).
%

\section{Templates}
In \Tabref{tab:temp}, we summarize the templates used for our synthetic data.

\begin{table}[h]
    \centering
    \small{
    \begin{tabular}{l}
    \toprule
         Insertion:\\
        
         \quad I want \textbf{xx}, I prefer \textbf{xx}, I like \textbf{xx}\\
         \quad I would like to see \textbf{xx}, Show me \textbf{xx},\\
         \quad Give me \textbf{xx}, I'm interested in \textbf{xx}\\
         \quad I need \textbf{xx}, Search for \textbf{xx}, Return \textbf{xx} \\
        (\textbf{xx} are nodes to be inserted)\\
         \midrule
         Deletion:\\
         \quad remove \textbf{xx}, I do not want \textbf{xx}, delete \textbf{xx} \\
         \quad I do not like \textbf{xx}, omit \textbf{xx}, I do not need \textbf{xx} \\
         \quad erase \textbf{xx}, ignore \textbf{xx}, discard \textbf{xx}, drop \textbf{xx}\\
         (\textbf{xx} denotes the node to be deleted)\\
        \midrule
        Substitution:\\
        \quad change \textbf{xx} to \textbf{yy}, update \textbf{xx} to \textbf{yy}\\
        \quad replace \textbf{xx} with \textbf{yy}, substitute \textbf{yy} for \textbf{xx}\\
        \quad I prefer \textbf{yy} to \textbf{xx}, modify \textbf{xx} to \textbf{yy}\\
        \quad I want \textbf{yy} rather than \textbf{xx}, switch \textbf{xx} to \textbf{yy}\\
        \quad convert \textbf{xx} to \textbf{yy}, give me \textbf{yy} instead of \textbf{xx}\\
        (\textbf{xx} and \textbf{yy} are old nodes and updated nodes)\\
        \bottomrule
    \end{tabular}
        }
    \caption{Simplified templates for synthetic data, with each operation has 10 templates.}
    \label{tab:temp}
\end{table}

\section{Examples from User-generated Dataset}
\begin{figure*}[ht]
\centering
  \begin{subfigure}[b]{0.4\linewidth}
    \includegraphics[width=\linewidth]{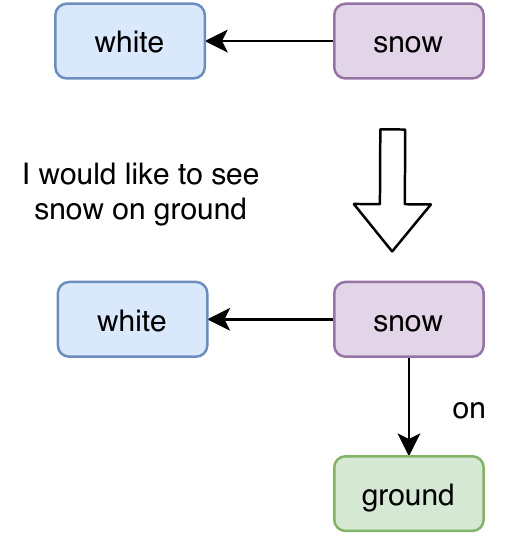}
     \caption{Insertion}
  \end{subfigure}\hfill
  \vspace{5mm}
  \begin{subfigure}[b]{0.4\linewidth}
    \includegraphics[width=\linewidth]{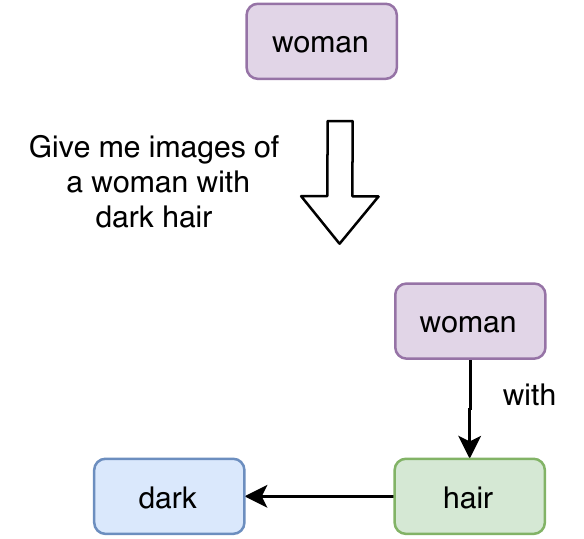}
    \caption{Insertion}
  \end{subfigure}
  \vspace{5mm}
  \begin{subfigure}[b]{0.4\linewidth}
    \includegraphics[width=\linewidth]{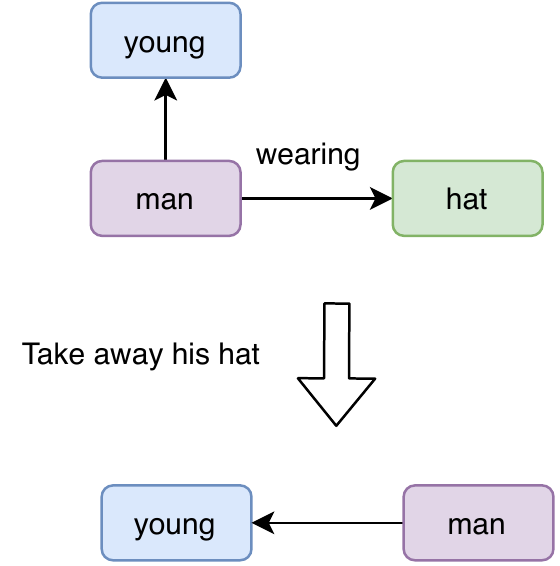}
    \caption{Deletion}
  \end{subfigure}
  \hfill
  \begin{subfigure}[b]{0.4\linewidth}
    \includegraphics[width=\linewidth]{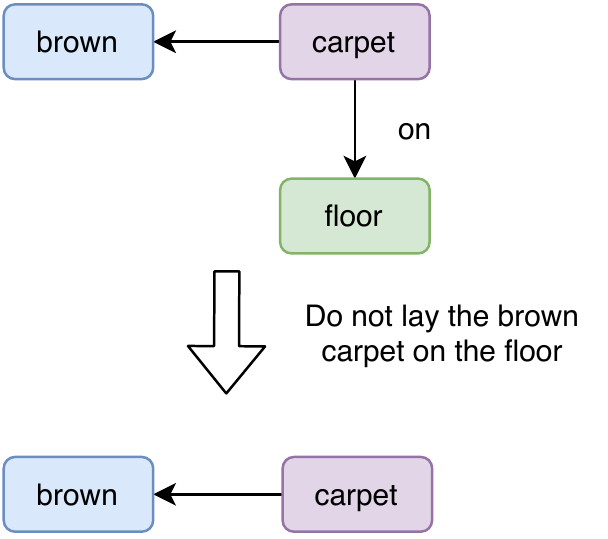}
    \caption{Deletion}
  \end{subfigure}
  \caption{Examples from the user-generated dataset.}
  \label{fig:examples}
\end{figure*}

We provides some examples of our user-generated dataset in \Figref{fig:examples}.

\section{Alignments between Different Components}
\begin{figure*}[h]
\centering
\begin{subfigure}[b]{0.4\linewidth}
    \includegraphics[width=\linewidth]{new_figures/e3.pdf}
    \caption{An example of graph modification}
  \end{subfigure}
  \hfill
  \vspace{5mm}
  \begin{subfigure}[b]{0.45\linewidth}
    \includegraphics[width=\linewidth]{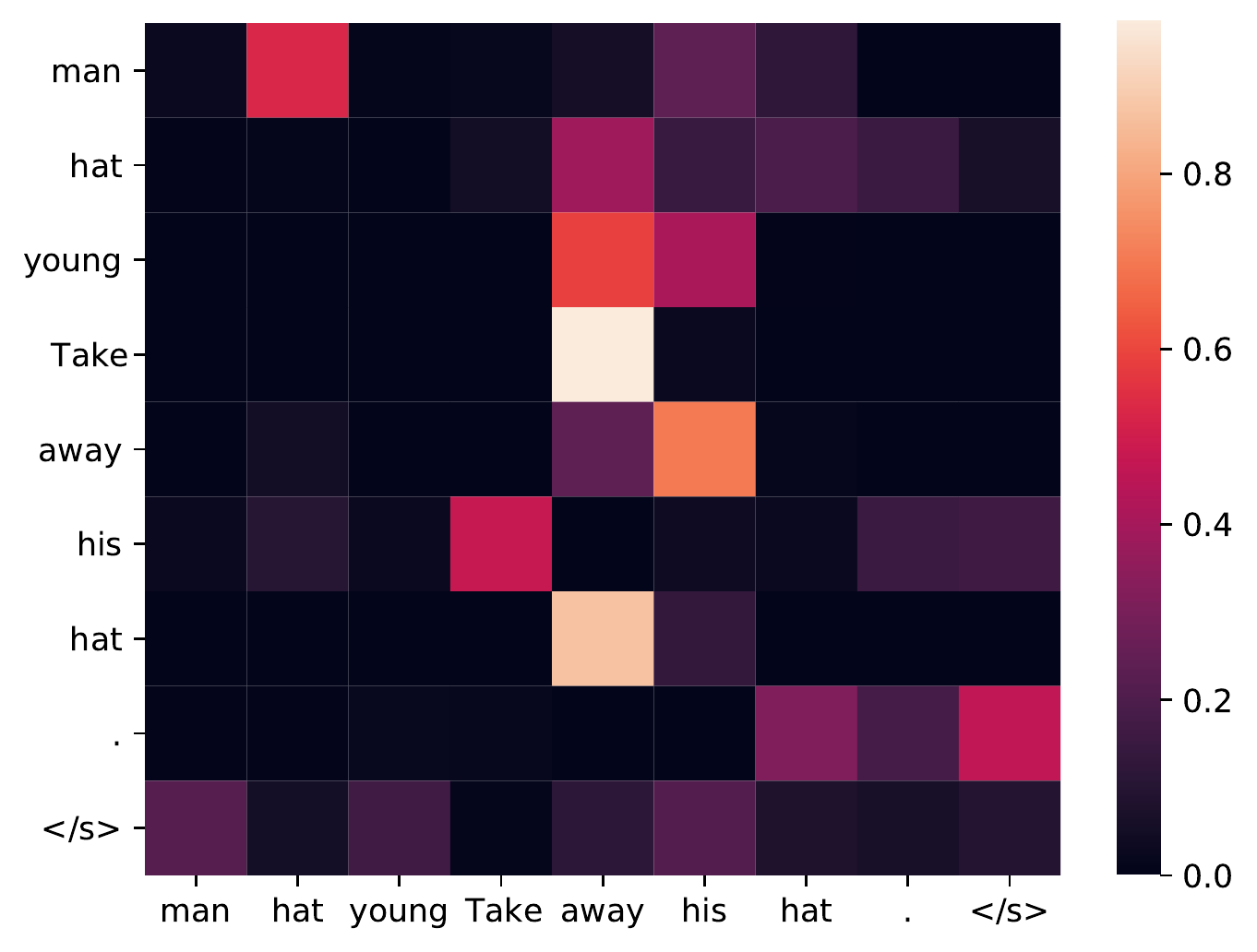}
     \caption{Cross Attention between query and source graph}
  \end{subfigure}\hfill
  \begin{subfigure}[b]{0.4\linewidth}
    \includegraphics[width=\linewidth]{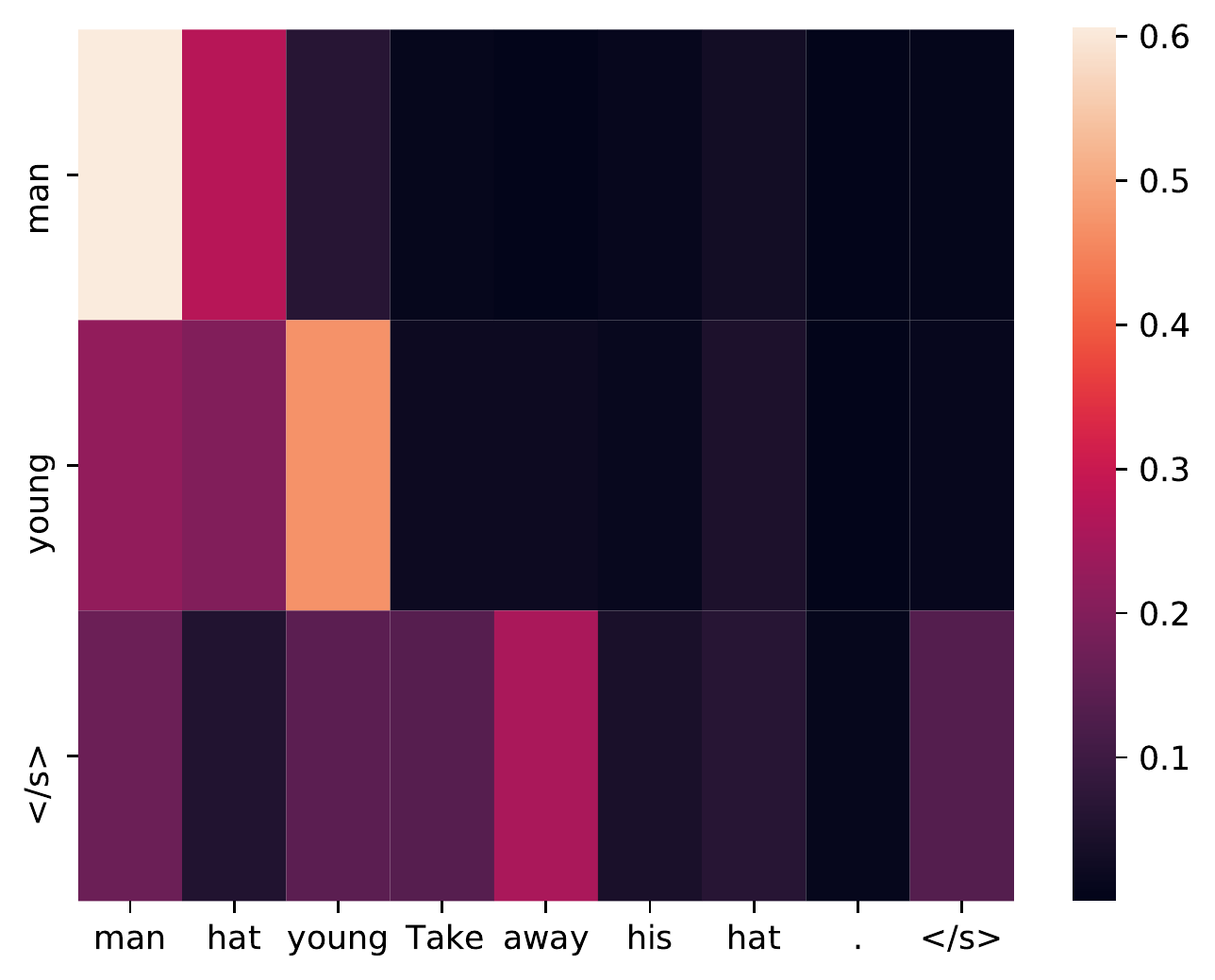}
    \caption{Attention between source information and target nodes}
  \end{subfigure}
  \hspace{10mm}
  \vspace{5mm}
  \begin{subfigure}[b]{0.4\linewidth}
    \includegraphics[width=\linewidth]{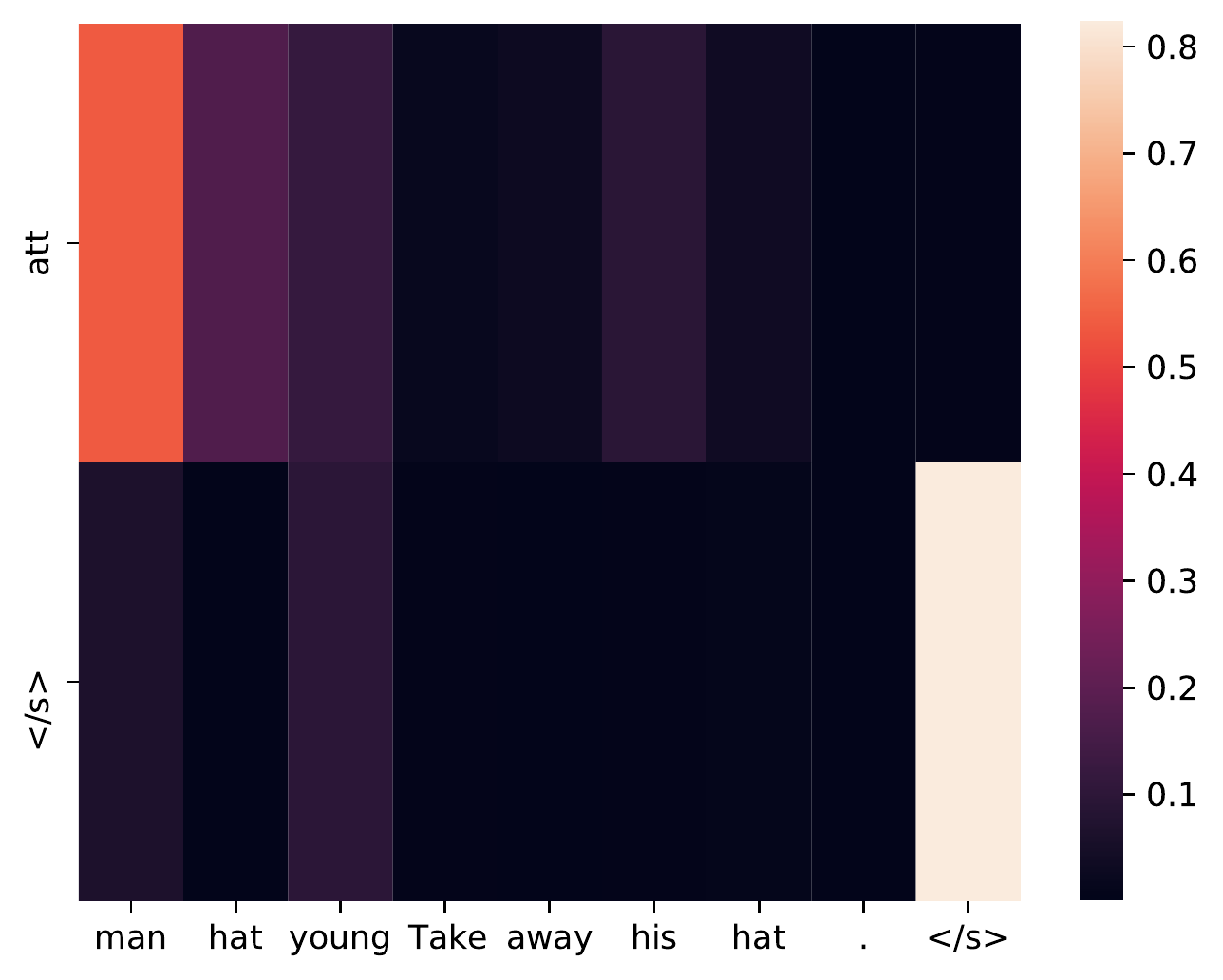}
    \caption{Attention between source information and target edges}
  \end{subfigure}
  \caption{Alignments between different components.}
  \label{fig:heatmap_e1}
\end{figure*}

\begin{figure*}[h]
\centering
\begin{subfigure}[b]{0.4\linewidth}
    \includegraphics[width=\linewidth]{new_figures/e1.pdf}
    \caption{An example of graph modification}
  \end{subfigure}
  \hfill
  \vspace{5mm}
  \begin{subfigure}[b]{0.45\linewidth}
    \includegraphics[width=\linewidth]{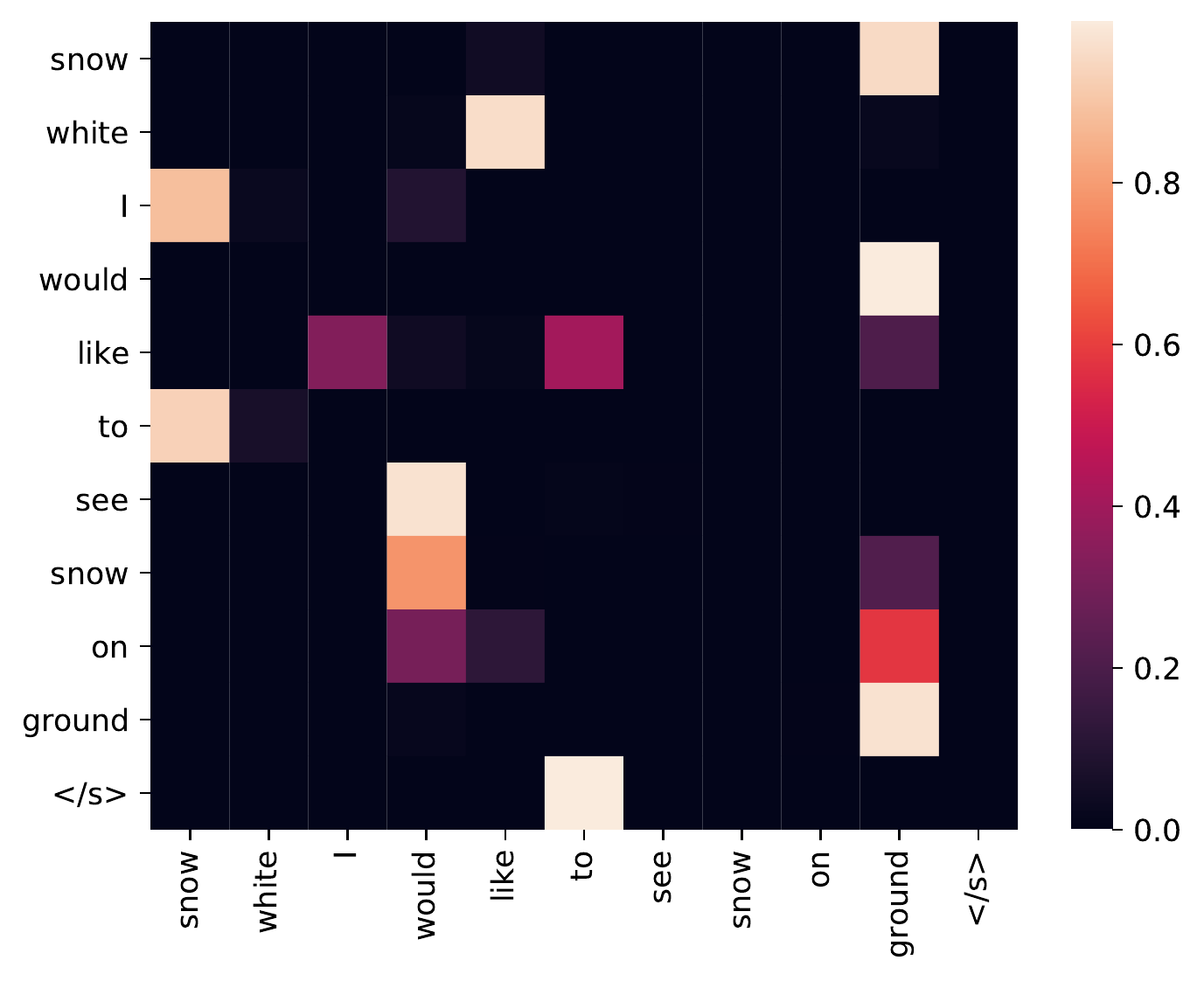}
     \caption{Cross Attention between query and source graph}
  \end{subfigure}\hfill
  \vspace{5mm}
  \begin{subfigure}[b]{0.4\linewidth}
    \includegraphics[width=\linewidth]{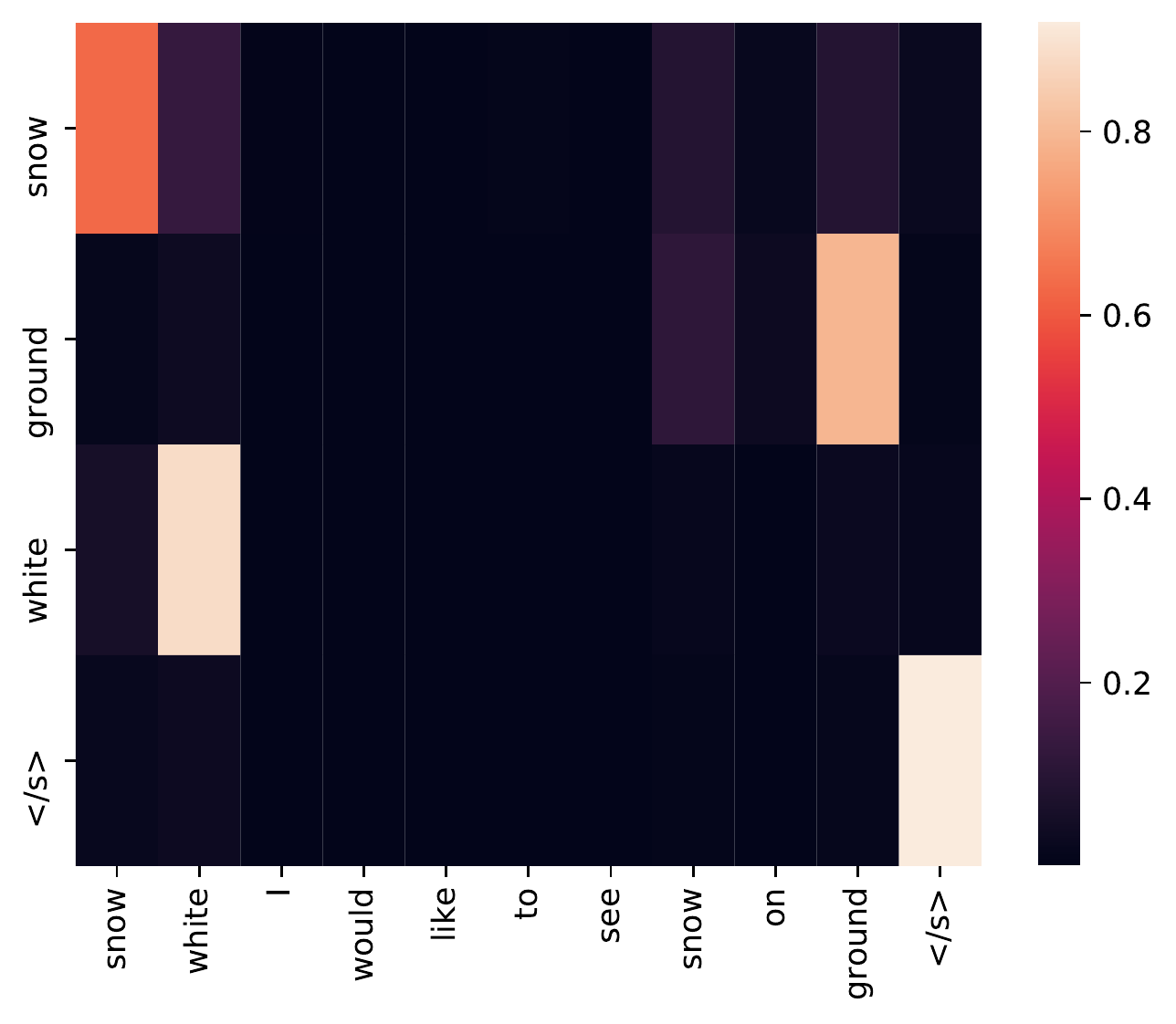}
    \caption{Attention between source information and target nodes}
  \end{subfigure}
  \hspace{10mm}
  \vspace{5mm}
  \begin{subfigure}[b]{0.4\linewidth}
    \includegraphics[width=\linewidth]{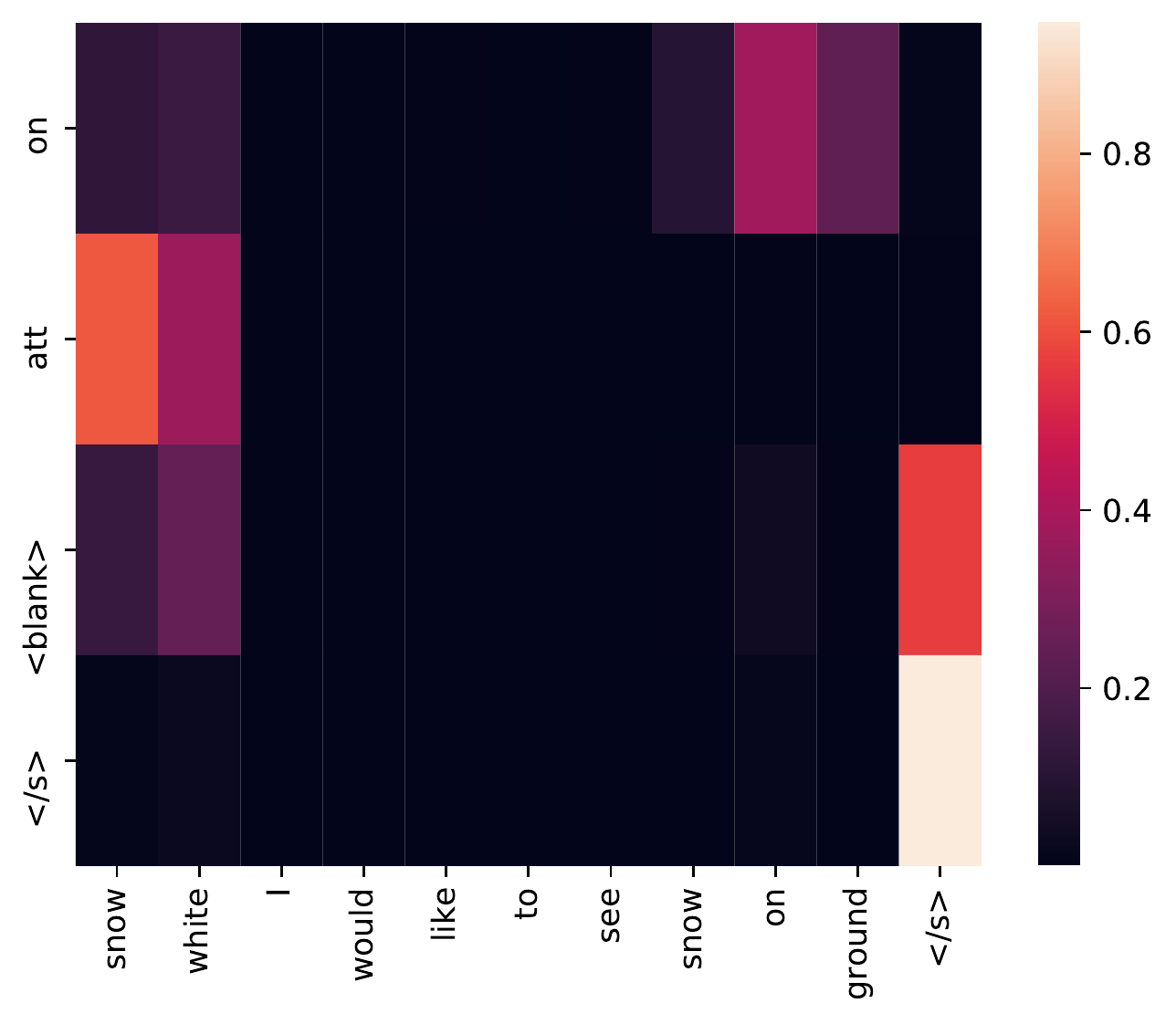}
    \caption{Attention between source information and target edges}
  \end{subfigure}
  \caption{Alignments between different components.}
  \label{fig:heatmap_e2}
\end{figure*}

\Figref{fig:heatmap_e1} and \Figref{fig:heatmap_e2} provide the alignments between different components of our cross attention model. Indeed, the cross attention is capable of aligning the source graph to the modification query.

\begin{figure*}[ht]
\centering
  \begin{subfigure}[b]{0.4\linewidth}
    \includegraphics[width=\linewidth]{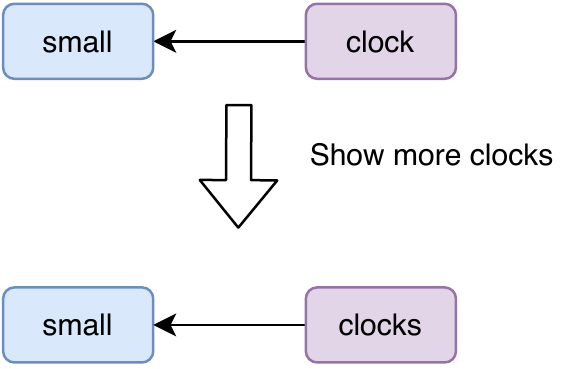}
     \caption{Object substitution}
  \end{subfigure}\hfill
  \vspace{5mm}
  \begin{subfigure}[b]{0.4\linewidth}
    \includegraphics[width=\linewidth]{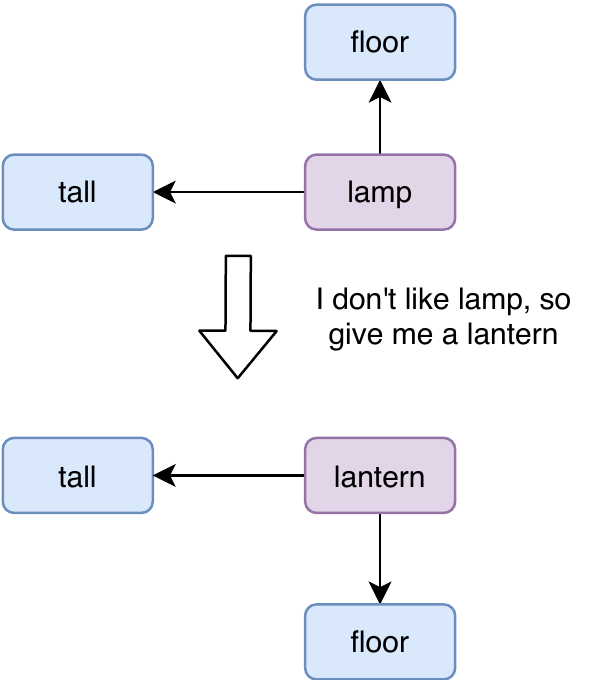}
    \caption{Object substitution}
  \end{subfigure}
  \vspace{5mm}
  \begin{subfigure}[b]{0.4\linewidth}
    \includegraphics[width=\linewidth]{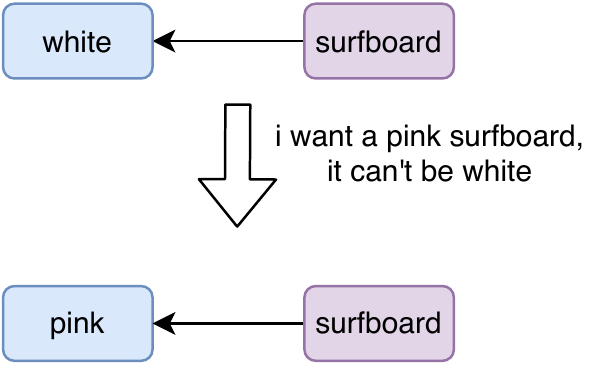}
    \caption{Attribute substitution}
  \end{subfigure}
  \hfill
  \begin{subfigure}[b]{0.4\linewidth}
    \includegraphics[width=\linewidth]{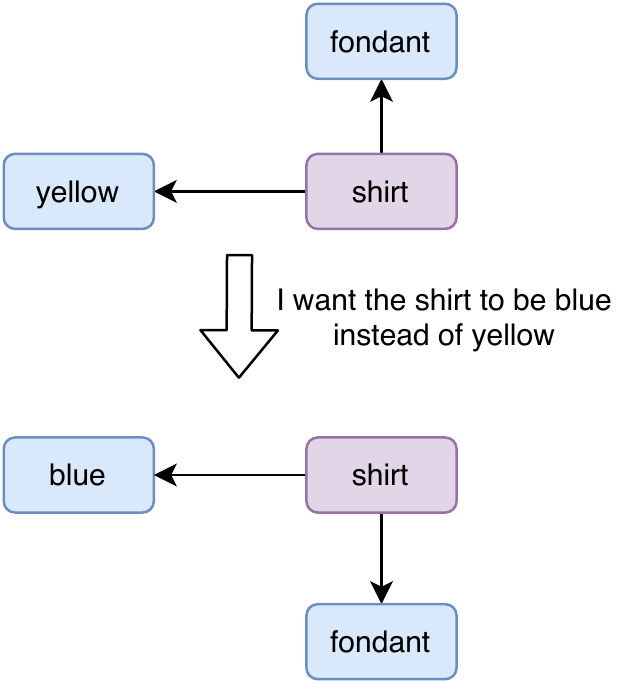}
    \caption{Attribute substitution}
  \end{subfigure}
  \caption{Examples from the user-generated dataset.}
  \label{fig:examples_sub}
\end{figure*}

\end{document}